\documentclass{article}

\usepackage[final]{neurips_2025}

\usepackage[utf8]{inputenc} 
\usepackage[T1]{fontenc}    
\usepackage{hyperref}   
\usepackage{multirow}
\usepackage{diagbox}
\usepackage{tikz}
\usepackage{url}            
\usepackage{booktabs}       
\usepackage{amsmath}
\usepackage{amsfonts}  
\usepackage{amsthm}
\usepackage{amssymb}
\usepackage{bm} 
\usepackage{subcaption}
\usepackage[capitalize,noabbrev]{cleveref}
\usepackage{nicefrac}       
\usepackage{microtype}      
\usepackage{xcolor} 
\usepackage{algorithm}
\usepackage{algorithmic}
\usepackage{graphicx} 
\usepackage{tabulary}
\usepackage{colortbl}

\newtheorem{theorem}{Theorem}[section]

\newtheorem{lemma}[theorem]{Lemma}

\newtheorem{definition}[theorem]{Definition}

\newtheorem{remark}[theorem]{Remark}

  \newtheorem*{replemma}{Lemma}

\newcommand{\shahrzad}[1]{\textcolor{red}{\textsc{Shahrzad:} \emph{#1}}}

\DeclareFontFamily{U}{stix2bb}{}
\DeclareFontShape{U}{stix2bb}{m}{n} {<-> stix2-mathbb}{}

\NewDocumentCommand{\indicator}{}{\text{\usefont{U}{stix2bb}{m}{n}1}}

\newcommand{\BuCC}{\textsc{BuCChoi}}
\newcommand{\PRIM}{\textsc{PRIM}}
\newcommand{\RIM}{\textsc{RIM}}
\newcommand{\DeepCheep}{\textsc{DyPChiP}}
\newcommand{\algOPTCard}{\textsc{CustOPT}}

\newcommand{\algPRIMabb}{\textsc{Profile-based \RIM (\PRIM)}}

\newcommand{\algSAMPLEabb}{\textsc{\topkgmm Sampling (\topKGMM)}}

\newcommand{\algPROFILEprob}{\textsc{Profile Probability}}

\newcommand{\primpos}{\textsc{primpos}}
\newcommand{\FindTop}{\textsc{FindTop}}

\newcommand{\topKGMM}{TopKGMM}
\newcommand{\topkgmm}{\topKGMM}
\newcommand{\GMM}{GMM}
\newcommand{\gmm}{\GMM}
\newcommand{\topKMM}{TopKMM}
\newcommand{\topkmm}{\topKMM}

\newcommand{\profile}[2]{\extt{#1}{#2}}
\newcommand{\pss}{\profile{S}{\sigma}}
\newcommand{\extt}[2]{T_{#2}({#1})}
\newcommand{\To}[1]{T_{#1}}

\newcommand{\K}{\mathcal{K} }
\newcommand{\Null}{\varnothing }
\newcommand{\choice}{\mathcal C }
\newcommand{\R}{{\mathcal R}}
\newcommand{\Rm}{{\mathcal R}^{\mathsf{M}}}
\newcommand{\D}{{\cD}}

\newcommand{\A}{{\mathcal A}}
\newcommand{\B}{{\mathcal B}}
\newcommand{\X}{{\mathcal X}}

\newcommand{\abs}[1]{\left|#1\right|}

\newcommand{\cA}{\ensuremath{\mathcal{A}}}
\newcommand{\cAN}{\ensuremath{\mathcal{A}^\Null}}

\newcommand{\cD}{\ensuremath{\mathcal D}}

\newcommand{\cI}{\ensuremath{\mathcal I}}

\newcommand{\RRN}{\mathbb{R}_{\ge 0}}

\newcommand{\probs}[2]{\ensuremath{\mathbb{P}_{#2}\left[#1\right]}}

\title{Generalized Top-$k$ Mallows Model for Ranked Choices}

\author{
  Shahrzad Haddadan \\
 Rutgers Business School  \\
 Piscataway, NJ 
 \\
   \texttt{shaddadan@business.rutgers.edu} \\
 \And
  Sara Ahmadian \\
Google research \\
Seattle, WA \\
sahmadian@gmail.com
}

\begin{document}

\maketitle

\begin{abstract}

 The classic Mallows model is a foundational tool for modeling user preferences. However, it has limitations in capturing real-world scenarios, where users often focus only on a limited set of preferred items and are indifferent to the rest. To address this, extensions such as the top-$k$ Mallows model have been proposed, aligning better with practical applications.
In this paper, we address several challenges related to the generalized top-$k$ Mallows model, with a focus on analyzing buyer choices. Our key contributions are: (1) a novel sampling scheme tailored to generalized top-$k$ Mallows models, (2) an efficient algorithm for computing choice probabilities under this model, and (3) an active learning algorithm for estimating the model parameters from observed choice data. These contributions provide new tools for analysis and prediction in critical decision-making scenarios.
We present a rigorous mathematical analysis
for the performance of our algorithms. 
Furthermore, through extensive experiments on synthetic data and real-world data, we demonstrate the scalability and accuracy of our proposed methods, and we compare the predictive power of Mallows model for top-$k$ lists compared to the simpler Multinomial Logit model.





\end{abstract}

\section{Introduction}
User preferences over a set of alternative items play a crucial role in various decision-making scenarios. A key concept in this context is the \emph{choices} customers make when presented with a subset of alternatives, referred to as an \emph{assortment}, drawn from a larger pool of items.  
The mathematical modeling of preferences and choices is both essential and challenging. It enables researchers and business leaders to analyze and predict customer behavior, thereby informing more effective decision-making. Probabilistic models over rankings, such as the Plackett-Luce (PL) and Mallows models, are widely used to represent user preferences. Based on the Plackett-Luce (PL) model, the Multinomial Logit (MNL) model has been suggested for modeling choice, and it has been extensively applied in choice modeling due to its simplicity and interpretability.

The Mallows model \citep{mallows1957non} is a distance-based probability distribution defined over permutations and is reminiscent of the Gaussian distribution for scalar variables. It has been used successfully to model preferences, particularly where ranking data is available. Recent work has demonstrated the high predictive power of the Mallows model in modeling choices \citep{desir2023robust, desir2021mallows}, sparking follow-up research into revenue management and assortment optimization in this framework \cite{desir2021mallows, desir2023robust, feng2022mallows, Riger24newassMM}. 
 One key challenge in applying the Mallows model more broadly is that its classic definition is restricted to full permutations.

In many real-world scenarios, however, user preferences are often observed as \emph{top-$k$ lists} rather than full rankings. For example, many platforms display only a subset of items to customers and ask them to rank a fixed number of preferred items. In recommender systems, advertising platforms, search engines, news aggregators, and social media friend suggestions, users are typically shown only the top-$k$ most relevant items, rather than an exhaustive list. Likewise, users often express preferences for a limited number of favorite items and show indifference toward the remainder.

This recurring structure in many applications has motivated the extension of the Mallows model to handle top-$k$ lists. The design of algorithm for this variant is significantly more complex than in the traditional permutation-based setting, leading to a growing body of research focused on learning, inference, and aggregation under the top-$k$ Mallows model \citep{lu2011learning, chierichetti2018mallows, collas2021concentric, vitelli2018probabilistic, fotakis2021aggregating, pmlr-v202-boehmer23a, goibert2023robust, 10.1145/3637528.3672014, qian2019weighted, akbari2022top}.

In this paper, we propose employing a generalized Mallows model for top-$k$ lists for choice modeling, and we develop efficient algorithms related to generating samples from it, learning its parameters and finding choice probabilities. 
Our approach addresses more realistic preference structures, where users are unlikely to hold complete rankings over a large set of items, an assumption that is often impractical in real-world applications.

\subsection{Related  Work}\label{sec:related}


\vspace{-0.2cm}
In this section we present relevant related work on choice modeling and Mallows model.
\vspace{-0.25cm}
\paragraph{Choice modeling}  Various probabilistic models have been developed to capture choice behavior, with the Multinomial Logit (MNL) \citep{bradley1952rank} model being the most widely used due to its simplicity and interpretability. In the MNL model  each product is assigned a positive score or weight, and the probability of selecting a product from an assortment is proportional to its score. Importantly, the MNL model satisfies Luce’s Choice Axiom, also known as the Independence of Irrelevant Alternatives (IIA).
While this property makes the MNL model analytically convenient, it also limits its expressiveness in capturing more complex choice behaviors.

To overcome the limitations of MNL, the mixture MNL model (also known as mixed logit) was introduced and popularized by \cite{mcfadden2000mixed}. Learning the parameters of a mixed MNL model from observed choices—where each observation is an assortment and a chosen item—is a key challenge. Early approaches used heuristics based on maximum likelihood estimation \citep{dempster1977maximum}, and more recent work has provided statistically rigorous methods with provable sample complexity guarantees \citep{pmlr-v80-chierichetti18a,NIPS2014_b33336b8,10.1145/3488560.3498506}.

Beyond MNL-based models, other frameworks such as the Mallows-based choice model \citep{desir2021mallows} and Markov chain-based models \citep{blanchet2016markov} have been proposed. These models offer greater flexibility but come with added complexity: tasks that are straightforward in MNL, such as computing choice probabilities or generating samples become substantially more challenging.
\vspace{-0.25cm}
\paragraph{Mallows model} The Mallows model (MM), originally introduced by \citet{mallows1957non}, defines a distance-based distribution over full rankings, where the probability of a permutation decays exponentially with its distance from a central (ground-truth) ranking. This property has made MM a useful foundation for preference modeling in machine learning. To better accommodate real-world data—where users often provide partial rather than full rankings—several extensions have been proposed. Early work by \citet{GMMVerduci1986} and \citet{JMLR:v9:lebanon08a} adapted MM to top-$k$ rankings. Later \citet{chierichetti2018mallows} proposed a new distance measure on top-$k$ lists and defined a parametrized Mallows model for top-$k$ lists based on it (TopKMM). 

One of the strengths of the standard Mallows model on permutations is the closed-form expression of its normalization constant which enables the design of several algorithms. 
For instance, the Repeated Insertion Method (RIM) \citep{RIMpaper} allows efficient sampling from the full-ranking Mallows model. However, this tractability does not extend to TopKMM: computing probabilities or generating samples becomes non-trivial due to the lack of a closed-form normalizing constant. While \citet{chierichetti2018mallows} proposed a dynamic programming approach to generate samples, they highlighted the absence of an RIM-like  sampling method for top-$k$ lists as an open problem.

\citet{desir2021mallows} proposed leveraging the Mallows model to capture choice behavior, demonstrating improved predictive accuracy over traditional models like the Multinomial Logit (MNL). However, a key challenge in applying the Mallows model to choice modeling lies in computing choice probabilities—i.e., the likelihood that a given item is selected from an assortment. Unlike MNL, where such probabilities have closed-form solutions the Mallows model does not admit such tractable computation in general. 
To address this, some works have introduced Mallows-like models designed to simplify  choice probabilities calculations \citep{feng2022mallows}. In contrast, \citet{desir2023robust} tackle the original model directly and develop a dynamic programming approach for computing choice probabilities by leveraging key ideas form the Repeated Insertion Method (RIM).


Learning the parameters of the Mallows model or its mixtures from full permutation, or top-$k$ list\footnote{We remark that some prior work considers scenarios where users select multiple items or a list of $k$ items—referred to as a top-$k$ choice. In contrast, our focus is solely on the selection of a single item, and we use the term \emph{top-$k$}
 only as a parameter of the Mallows model.}, samples has been extensively studied in prior work \citep{Liu2018EfficientlyLM, soda2008Mossel, Neurips14Awtashi, NEURIPS2020_6affee95, pmlr-v97-tang19a, collas2021concentric, akbari2022top,liu2018efficiently,Chiermixture2015}.
 Some studies focus on learning these parameters from historical data, while others consider an \emph{active learning} setting, where upon each consumer’s arrival, the platform adaptively selects the assortment of items to offer based on past observations \cite{susan2022active}.
In contrast, learning model parameters from partial observations presents a greater challenge. Several works address this problem by studying learning from pairwise comparisons \cite{lu2014effective,vitelli2018probabilistic,pmlr-v97-tang19a}. A less explored problem is estimating the model parameters when only the users’ choices from offered assortments of arbitrary sizes are observed. Existing methods in this setting often lack finite-sample complexity guarantees, limiting their theoretical robustness.

\subsection{Summary of Contributions}
In this paper, we focus on Mallows model on top-$k$ lists, and we  make several algorithmic contributions for the usage of this model.
In order to obtain our results in full generality, we extend \citet{chierichetti2018mallows}'s model on Mallows model for top-$k$ list to a generalized  version by associating a weight to each product. Our algorithmic contribution are listed as follows:


\vspace{-0.3 cm}
\begin{enumerate}
    \item  \emph{Sampling:} \textsc{Profile based Repeated Insertion Method} (\PRIM): 
The \emph{Repeated Insertion Method} (\RIM) is a common method for sampling permutations in the classic Mallows model. However, extending this method to the top-$k$ Mallows model remains an open problem. Our proposed algorithm, \PRIM, offers similar functionality to \RIM\ and improves upon the prior dynamic programming approach by \cite{chierichetti2018mallows}, reducing the runtime from  $O(k^24^k+k^2\log n)$ to $O(k2^k+k^2\log n)$~.

    
    \item \emph{Choice Probabilities:} \textsc{DYnamic Programming for CHoIce Probabilities (\DeepCheep)}: \DeepCheep\ is an algorithm to calculate the choice probabilities when they are inferred from a Mallows model on top-$k$ lists.    This result extends the work of  \cite{desir2021mallows} who consider the classic Mallows model for permutations.
    \item \emph{Learning the Center:} \textsc{BUild Center from CHOIces} (\BuCC): \BuCC\ is an \emph{active learning algorithm} designed to learn the center of a top-$k$
 Mallows model distribution. It operates by presenting assortments of a specified size $r$ to customers and, based on their observed choices, infers both the ranking of the center and the size of the center $k$.

    
\end{enumerate}
   The accuracy and complexity of these algorithms are demonstrated through rigorous mathematical analysis as well as  experiments on real-world and synthetic data. \\
Furthermore, we apply these algorithms   and fit a top-$k$ Mallows model to  a  real-world publicly available dataset including users' preferences over 100 sushi types, represented as top-10 lists \citep{sushidataset2005}. This model helps us  predict choice probabilities with high accuracy, and our results  demonstrate that the top-$k$ Mallows model achieves significantly higher predictive accuracy than the Multinomial Logit model on this dataset




\vspace{-0.2cm}
\section{Preliminaries and Definitions}
\vspace{-0.2cm}

Let $N=[n] := \{1,2,\cdots, n\}$ represent a universe of $n$ elements. A top-$k$ list is a partial order on $N$ structured as 
$i_1 \succ i_2 \succ \cdots \succ i_k \succ \{i_{k+1}, \cdots, i_n\}$ , where the top-$k$ elements are strictly ordered, while the remaining $n-k$ elements are incomparable to each other. The collection of all top-$k$ lists over $N$ is represented by $T_{k,N}$, where $T_{n,N} = S_N$  corresponds to the symmetric group on $N$. 

For a top-$k$ list $\tau$ and a position $l\in [k]$, $\tau(l)$ refers to the element ranked at position $l$, while $\bar\tau$ denotes the set of elements ranked below the top-$k$. For simplicity, we sometimes use $\tau$ to represent the top-$k$ elements $\{\tau(1),\tau(2),\dots \tau(k)\}$. Thus, $i \in \tau$ indicates that $i$ is ranked among the top-$k$ elements of $\tau$, and set operations like $\subseteq$ and $\cap$ are applied accordingly. For $i,j\in N$, $i\succ_\tau j$ means $i$ is ranked above $j$ in $\tau$, i.e., $i\in \tau$ and either $j\in \bar\tau$ or $j\in \tau$ but ranked below $i$. Additionally, $i \parallel_\tau j$ indicates $i$ and $j$ are incomparable (both are in $\bar\tau$), while $i\perp_\tau j$ means they are comparable ($i\succ_\tau j$ or $j \succ_\tau i$).  

In this paper, we utilize the widely recognized Kendall's Tau distance, a commonly used metric that quantifies the number of pairwise disagreements between two permutations \citep{fagin2003comparing, critchlow2012metric}. This concept has been extended to top-$k$ lists, where it no longer forms a true metric but retains useful mathematical properties. 
Given a parameter $p\geq 0$, the $p$-parametrized distance between $\tau, \tau' \in T_{k,N}$ is defined as 
\begin{equation}\label{eq:topInversion}
\K^p(\tau,\tau') = \sum_{i,j \in \tau\cup \tau': i < j} \K^p_{i,j}(\tau,\tau') \text{, where} ~   \K^p_{i,j}(\tau,\tau') = \begin{cases}
1 & \text{ if } (i \succ_{\tau'}  j ~\&~ j \succ_{\tau} i) \text{ or vice-versa} \\ p & \text{ if } 
(i \perp_{\tau'} j ~\&~ i \parallel_{\tau} j) \text{ or vice-versa}\\ 0 & \text{ otherwise.}  \end{cases} 
\end{equation}

Using this distance measure, \citet{chierichetti2018mallows} define the {\em Mallows model for the top-$k$ lists}. Given a center $\tau^*$ and a decay parameter $\beta$, the probability distribution $\cD$ over top-$k$ lists $T_{k,N}$ is defined as:
\begin{equation}\label{eq:topkMM}\tag{\topKMM}
     \probs{\tau \in T_{k,N}}{\cD} \propto \exp\left(-\beta\  \K^p(\tau,\tau^*)\right) ~.
\end{equation}

To simplify our notation, we assume, without loss of generality, that the center 
$\tau^* $ is always the identity list $ 1 \succ 2 \succ \cdots k \succ \{k+1, \cdots, n\}$. 
Therefore, we denote $\K^p(\tau,\tau^*)$ as $\K^p(\tau)$. For full rankings, where all elements are comparable, we simply use $\K(\tau)$.

A natural extension of this model arises when the elements have associated weights. {\em Generalized Mallows Model} (\GMM) \cite{GMMVerduci1986} considers this case for full rankings: Given a decay parameter $\beta$ and non-negative weights $w_i \in \mathbb{R}_{\geq 0}$ for each $i\in N$, the probability distribution $\cD$ over rankings is defined as:  
\begin{equation}\label{eq:GMM} \tag{GMM}
     \probs{\tau \in T_{n,N}}{\cD} \propto \exp\left(-\beta\ \sum_{i,j: i< j} w_i \K_{i,j}(\tau)\right)
\end{equation}
where $\K_{i,j}(\tau)$ is 1 iff $\tau$ disagrees with $\tau^*$, i.e., $\tau$ ranks $j$ before $i$ for $j > i$ and 0 otherwise, as defined in \eqref{eq:topInversion}. In this model, each disagreement contributes the weight of the item ranked higher in $\tau^*$. This formulation can be simplified by the use of inversion vectors as follows: 
\[
\probs{\tau \in T_{n,N}}{\cD} \propto \exp\left(-\beta\sum_{i\in [k]}\ w_i I_i(\tau)\right) \text{ where $I_i(\tau) = \sum_{j: j>i} \K_{i,j}(\tau) = \sum_{j:j>i} \indicator\left(j \succ_\tau i \right)$}
\]
where $\indicator$ is an indicator function that takes the value 1 when true and 0 otherwise.

The generalized mallows model has traditionally been defined only for full rankings, and we extend it to {\em Generalized Mallows Model for Top-$k$ lists}. 
In this setting, each element $j\in \tau^*$ is assigned a non-negative weight, i.e., $w_i \in \mathbb{R}_{\geq 0}$ for $i\in [k]$, along with an additional weight $w_0\in \mathbb{R}_{\geq 0}$ for any element in $\bar\tau^*$. We use $\bm{w} \in {\mathbb{R}^{k+1}_{\geq 0}}$ to represent this collection of weights, which uses the following extension for inversion vectors: 

\begin{definition}[Inversion Vectors of a Top-$k$ list]\label{def:inversiontable} Given a top-$k$ list  $\tau\in T_{k,N}$, there are three components for inversion vectors: vectors $\mathbf{I(\tau), P(\tau)} \in \RRN^k$ where for $i \in [k]$:  
\begin{equation*}
    I_i(\tau)=\sum_{j: j>i} \indicator\left(j\succ_\tau i  \right), 
P_i(\tau)=\sum_{j: j>i} \indicator\left(i,j \in \bar\tau\right)
\end{equation*} 
and $Q(\tau) = {{k - \ell}\choose {2}}$ where $\ell=|\tau \cap \tau^*|$.
\end{definition}

Note that this definition is an alternative way to count the disagreement between $\tau$ and $\tau^*$ where the disagreement is always assigned to the higher-ranked element by $\tau^*$. When neither element is ranked higher ($i,j \notin \tau^*$), disagreements are assigned to $Q$. For example, for $\tau=(2,1,6,5)\in \To{4,[8]}$, we have $\mathbf{I}(\tau)=[1,0,2,2]$, $\mathbf{P}(\tau)=[0,0,1,0]$, $Q(\tau) = 1$. $I_1(\tau) = 1$ since element $2$ is ranked higher than $1$ by $\tau$, $I_3(\tau) = 2$ since elements $5$ and $6$ are ranked higher than $3$ and the same argument applies to $I_4(\tau)$. $P_3(\tau)=1$ since elements $3$ and $4$ are not comparable by $\tau$ (but they are ranked by $\tau^*$) and this disagreement is assigned to element $3$ as it is ranked higher than element $4$ by $\tau^*$. $Q$ counts disagreements for elements not ranked by $\tau^*$, i.e., elements $5$ and $6$. 

\paragraph{Generalized Mallows Model for Top-$k$ lists}
Given the center $\tau^*\in \To{k,N}$ and parameters $\beta\geq 0$, $\bm{w}\in \mathbb{R}^{k+1}_{\geq 0}$ and $p>0$, the probability distribution $\cD$ over $\To{k,N}$ is defined as:
\begin{equation}\label{eq:topkGMM}\tag{TopKGMM}
\probs{\tau\in \To{k,N}}{\cD}\propto \exp\left(-\beta \ {\K}^{p,w}(\tau,\tau^*)\right)
\end{equation}
where  
\begin{equation*}
  {\K}^{p,w}(\tau,\tau^*) := w_0 p Q(\tau) + \sum_{i\in [k]} w_i \cdot(I_i(\tau)+pP_i(\tau)). 
\end{equation*}

Note that setting $\bm{w}=\bm{1}$, we obtain \cref{eq:topkMM}, and 
$k=n$, recovers \cref{eq:GMM}.

\vspace{-0.2cm}\section{Sampling from \topkgmm}\label{sec:gen_mallows}
\vspace{-0.2cm}

We begin by investigating the challenge of efficiently sampling from \topkgmm. Although \citet{chierichetti2018mallows}'s  sampling algorithms for \topkmm\ can easily be extended to incorporate weights as in \topkgmm, here our concentration is to develop a sampling algorithm with similar functionality to the \emph{repeated insertion model}, which was left open \citep{chierichetti2018mallows}. The theory we develop here not only helps in the design of the sampling algorithm \PRIM\ but also plays a crucial role in the development of \DeepCheep, our proposed algorithm for choice probabilities.



\begin{theorem}[Sampling from \topKGMM]\label{thm:sampling} For a given instance $\cD$ of \topKGMM\,  
there exists an algorithm that efficiently samples a top-$k$ list according to $\cD$ in time complexity $O(k\cdot 2^\gamma + k \log n )$,\footnote{Here $O(k2^\gamma)$ is the complexity of a pre-processing step, after that each sample can be generate at cost $O(k \log n)$. } and space complexity $O(k\cdot 2^\gamma)$; where $\gamma=\min\{k,n-k\}$. 
\end{theorem}



Similar to the RIM method for permutations \citep{RIMpaper}, the core idea of our approach involves iteratively adding elements to a partially ordered sequence until exactly $k$ elements are sampled. While this strategy results in a correct sampling scheme in the context of permutations, for  top-$k$ lists it is essential to partition the sample space based on several features before iterative insertions begin. Our careful definition of inversion vectors in \cref{eq:topInversion} plays a crucial role here, as it allows us to focus on the behavior of elements in $[k]$ and specifically those sampled by $\tau$. 
This insights leads us to introduce {\em profiles} which represent the shared top-$k$ elements $\tau$ and $\tau^*$.
\subsection{Profile-based \topkgmm\ Distribution}\label{sec:profile}

Let profile $ S \subseteq [k]$ represent the subset of sample top-$k$ elements for a given top-$k$ list $\tau$. Formally,

\begin{definition}
[top-k Profile]\label{def:profile}
Given a center $\tau^*\in \To{k,N}$ with corresponding \topKGMM~ distribution $\cD$, we call a set $S\subseteq \tau^*$a {\em profile} and we define  $\extt{S}{\tau^*}$ and its probability with respect to  \D\ as:
\begin{equation}\label{eq:profile}
    T_{\tau^*}(S)=\{\tau{\mid} \tau\cap\tau^*{=}S\}, \quad 
    \probs{S}{\D} 
     =\sum_{\tau\in T_{\tau^*}(S)}     \probs{\tau}{\D} ~.
\end{equation}
When the center is clear from the context, we may simply use $T(S)$.
\end{definition}


 For any $\tau\in T(S)$, since $Q(\tau) = {k-|S| \choose 2}$ only depends on $S$, we  simply write $Q(S)$. The inversion vector $P$ counts the number of lower-priority elements that were previously ranked strictly lower but are now incomparable. An element $j$ has a positive $P_j(\tau)$ only if it is not ranked by $\tau$, in which case its value is given by $P_j(\tau) = n-k-(k-\ell)$ as exactly $k-\ell$ elements from $\{k+1, \cdots,n\}$ are now ranked by $\tau$. Since $P$ strictly depends on $S$, we can use $P(S)$ instead of $P(\tau)$. 
 
 The inversion vector $\mathbf{I}$ is defined for elements in $[k]$, and  for each element, is the number of lower-priority elements (w.r.t. the center) that are now ranked higher in $\tau$. For any element $j$ not ranked by $\tau$, i.e., $j \in [k]\setminus S$, $I_j(\tau) = |\{j+1, \cdots, k\} \cap S|$. Thus part of vector $I$, is entirely determined by $S$ and  independent of  $\tau$. Since $Q(S)$, $\mathbf{P}(S)$, and $\mathbf{I}_{j\in [k]\setminus S(\tau)}$ are independent of  rankings among $\tau$'s elements  and depend only on $S$, we  compute the probability of $S$ given this information. 
\begin{lemma}\label{lem:profile_prob}
Given a \topkgmm\ distribution $\D$ with  parameters $\beta$, $p$ and $\bm{w}$, any profile $S = \{s_1, s_2, \dots, s_\ell\} \subseteq [k]$ ($s_1<s_2<\cdots<s_l$) has probability $\probs{S}{\D}$ proportional to $\exp\left(-\beta f(S) \right)Z(S)$, where: 
\begin{equation*}
    f(S) = w_0 p Q(S) + \sum_{j \in [k]\setminus S} w_j (I_j(S)+p P_j(S))~, \text{and}\quad 
    Z(S) = {n-k \choose k-\ell} (k-\ell)! \prod_{j=1}^{\ell} \sum_{r=0}^{k-j} e^{-\beta w_{s_j} r} \enspace .
  \end{equation*}

and it can be sampled in $O(2^{\gamma} k)$; where $\gamma=\min\{k,n-k\}$.
\end{lemma}

\subsection{Sampling Algorithm}\label{sec:samplealg}
Building on the results from the previous section and leveraging the concept of a \emph{profile}, we now propose a method for  sampling from the \topkgmm\ distribution (\cref{alg:sample}). We first sample a profile $S$ with probability $\probs{S}{\D}$ (according to \Cref{lem:profile_prob}), and then we sample $\tau\in T(S)$ by inserting the elements $j\in S$ based on their contribution in inversion vectors  
\Cref{eq:topkGMM}). The later step can be 
viewed as generalizing the Repeated Insertion Method (\RIM). 
We refer to our method as \textsc{Profile-based-Repeated-Insertion-Method (PRIM)}. 

In \PRIM, we generate a top-$k$ list proportional to its sampling probability in $\cD$ conditioned on the fact that the common elements with the center top-$k$ ranking is $S$. Specifically, let $\ell = \left|S\right|$. We begin with an empty array $A$ and then sample $k-\ell$ elements from $[n]\setminus [k]$ 
 uniformly at random. Then we sequentially, insert the elements in $S$ in increasing order of their priority. When processing element $s$,  we insert $s$ in the current array $A$ at position $j \in \{0,1,\dots ,|A|\}$  with probability proportional to 
 \begin{equation}\label{eq:rimprob}
 Pr(\text{inserting $s$ at position $j$}) \propto \exp\left(-\beta w_{s}\cdot j\right)  
 \end{equation}
 as $j$ is the number of inversion associated with element $s$ when it is inserted at position $j$. Note that higher priority elements do not contribute to inversion of $s$ so when their position when they are inserted later is not important for $s$.
See \cref{app:profile_proofs} for  pseudocodes and proofs. 

\vspace{-0.2cm}
\section{Choice modeling and probabilities}
\vspace{-0.2cm}
In this section, we focus on the problems related to \emph{choice}. In particular we focus on the analysis of problems which help us predict the \emph{(top) choice} of a customer from a set of alternatives, a.k.a an \emph{assortment} using  \topkgmm. After presenting definitions we  design an algorithm which efficiently calculates choice probabilities having the distributions parameters. In \cref{sec:learning} we study the opposite problem in this context, which is learning the center of a  \topkgmm\ distribution by observing its  choice data. 


Given a set of products $[n]$, an \emph{assortment} is  any subset $\A$ of $[n]$. 
When $\A$ is offered to  a customer, she may \emph{choose} any of its elements or the ``no purchase'' option denoted by $\Null$. 
We use $N = [n] \cup \{\Null\}$ to denote all purchase options and correspondingly, we define $T_{k,N}$ to denote all top-$k$ lists over $N$.
We assume that the preferences of customers are derived from \topkgmm, and if we have multiple customer types, we use a mixture of several \topkgmm's where each may have different parameters. 

Given a top-$k$ list $\tau\in\To{k,N}$,
we define the \emph{choice} function $\choice_{\tau}:2^{[n]}\rightarrow N$ as follows: 
for any assortment $\A\subset [n]$, $\choice_\tau(\A)=i$, iff $i$ is the highest ranked element in $\A\cup \{\Null\}$ with respect to $\tau$. If all elements of $\A\cup \{\Null\}$ are incomparable w.r.t. $\tau$, then $i$ is taken uniformly from $\A\cup \{\Null\}$.

We focus on applying the \topkgmm\ model to represent customer preferences. Specifically, we assume that 
$\tau$ is sampled from a distribution 
$\cD$, where 
$\cD$
 is a \topkgmm\ distribution characterizing a single customer type. More generally, $\cD$  can represent several customer types and we may use a mixture distribution. Since choice probabilities of a mixture distribution  can simply be obtained as a linear combination of its singleton components, here we focus on singleton distributions.


\subsection{Calculation of Choice Probabilities: \DeepCheep}\label{sec:choice} 


Let  $\cD$ be a \topkgmm\ distribution on  $\To{k,N}$. We let  
 $\choice_\cD$ be a function mapping any assortment $\A \subseteq [n]$ and an option  $i\in \A\cup \{\Null\}$ to the probability that $\choice_\tau(\A)=i$ where $\tau$ is sampled from $\cD$. Formally,  $\choice_\cD: N\times 2^{[n]}\rightarrow [0,1]$ is defined as follows:
\begin{equation}\label{eq:choiceprob}
    \choice_{\cD}(i,\A)= \sum_{\tau\in \To{k,N}} \probs{\tau}{\cD} \cdot \indicator\left(C_\tau(\A)=i \right)~.
\end{equation}
We now introduce \DeepCheep\ which 
calculates \emph{choice probabilities} as defined in \cref{eq:choiceprob}, and its correctness and runtime complexity is stated in the following theorem:

\begin{theorem}[Calculation of choice probabilities]\label{thm:choicep}
Given an assortment $\cA\subseteq [n]$ and a \topKGMM\ instance 
$\cD$, 
  \DeepCheep\ calculates $\choice_{\D}(j,\cA)$ for all $j\in \cA\cup\{\Null\}$ in $O\left(2^{\min\{k,n-k\}} k^3 \left|{\A}\right|^2\right)$.   
\end{theorem}

The main idea of this algorithm is to find these probabilities by conditioning on a given profile. 
Our dynamic programming tables are defined based on the order in which items are considered in  \PRIM.

\paragraph{Overview of \DeepCheep}
Consider profile $S\subseteq [k]$ with $\ell=\left|S\right|$. Let $\cAN = \cA \cup \{\Null\}$. For a 
top-$k$ list $\tau\in T(S)$, item $a$ from $\cA$ is picked if (i) $\tau$ does not include any item from $\cAN$ and so $a$ is picked randomly from $\cA^{\Null}\subseteq \bar{\tau}$, or (ii) $a$ is  top-ranked in $\cAN$ w.r.t $\tau$. We handle the two cases separately: 

\begin{enumerate}
    \item
    Let $\bar{\pi}_S(a)$ denote the probability associated to case (i). Note that $\tau\cap \cA^{\Null}=\emptyset$ implies that $\tau$ is from a profile $S$ with $S\cap \cA=\emptyset$. Thus for any $a\in \cAN$: 
\begin{equation*}\label{eq:choiceprob:prep}
\bar{\pi}_S(a)=
\mathbf{P}\left(\tau\cap \cA^{\Null}=\emptyset\right)\cdot \frac{1}{\left|\cA \right|+1}=
\indicator\left(
\cA^{\Null}\cap S= \emptyset 
\right)
\cdot 
\frac{{n-k-(|\cA|+1) \choose{k-\ell}}}{{{n-k}\choose{k-\ell}}}\cdot \frac{1}{|\cA|+1}\enspace.
\end{equation*}
\item 
Let $\bar{\cA}\doteq \cAN\cap \bar{\tau}$. The elements in $\cAN$ who have a non-zero probability of being top ranked at some iteration of DP table are only in $\bar{\cA}\cup S$.  We calculate these probabilities by conditioning on two other parameters: (1) where in \PRIM\ algorithm they have a chance of being sampled, and (2) the position in which they are positioned in the top-$k$ list when they are the winner, i.e, ranked highest among $\cAN$. 
We use a  three dimensional dynamic programming table   $\pi_S$  as follows:
Let $\{a_1,a_2,\dots, a_r, s_\ell, s_{\ell-1},\dots, s_1\}$  be an ordering of the elements in $\bar{\cA}\cup S$ where the first segment  
is an arbitrary ordering of $\bar{\cA}$ and the second 
is the ordering of $S$ used in \PRIM. \\
For $q=0$, we let $\pi_S(i,j,q)$  be the probability that any element
$a_i\in \{a_1,a_2,\dots, a_r\}$ is ranked $j$th and higher than all other elements of $\bar{\cA}$ (see \cref{eq:choiceprob:firstcond}). 
Then, by iterating over $q=1,2,\dots \ell$, 
we consider  the   $q$th iteration of  the for loop in  \PRIM, and for any position $1 \leq j \leq q$, we define $\pi_S(i,j,q)$  be the probability that after completion of  the $q$th iteration of the for loop in \PRIM, $a_i$ is the highest element  in $\cAN$ (the winner) which has so far been sampled, and $a_i$ is so far ranked $j$-th. \\
We update these probabilities by conditioning whether the newly inserted  element, i.e., $s_{\ell+1-q}$ is added before the prior winner, or after it. 
Note that, since we the profile $S$ is fixed, the probability of inserting  a new element  in a particular location may be obtained from \cref{eq:rimprob}. For details of the recursive definition of $\pi_S(i,j,q)$ please see \cref{app:choice}.  
After the dynamic programming table is filled, we may calculate the choice probability of each $a_i\in \cAN$ conditioned on profile $S$ with $|S|=\ell$ by:
\[
\pi_S(a_i)=\sum_{j=1}^\ell
\pi_S(i,j,\ell)~.
\]
\end{enumerate}

In each cell of the table, we need to look at  $O(k)$ other cells which involves at most $O(k\left|\A\right|)$ operations so each table entry can be calculated in $O(k\abs{\A})$ time. Finally, we may calculate the 
\begin{equation}\label{eq:choiceprobsum}
\forall a\in \cA, ~
  \choice_\cD(a,\cA)= \sum_{
\substack{S\subseteq [k]
} } \probs{S}{\cD}\left(\pi_S(a)
+  \bar{\pi}_S(a)
\right) ~.
\end{equation}

\vspace{-0.4cm}
\paragraph{Runtime of \DeepCheep}
The size of the dynamic programming table is $\abs{\A}k^2$, and calculation of each element needs $ k\abs{\A}+n$ operations. 
In \cref{eq:choiceprobsum} we have a sum over all profiles; which is bounded by $O(2^{\min\{k,n-k\}})$. Thus, in total all choice probabilities may be calculated in time $\Theta(2^\gamma k^3\abs{\A}^2+2^\gamma k^2\abs{\A}n)$; $\gamma=\min\{k,n-k\}$.

\vspace{-0.2cm}
\section{Learning the center from choice data}\label{sec:learning}

In this section, we focus on the problem of learning the center of a \topkgmm\ distribution from \emph{choice data}. This task is more challenging than learning from  complete top-$k$
 samples, as each data point provides limited information and the choice data depends on the specific assortments presented.

Consider  a \topkgmm\ distribution $\cD$ on  $\To{k,N}$, the goal is to construct the center of $\cD$, namely $\tau^*$ by observing  choice data. The choice data consists of pairs $(\A_t, c_t)$, where $\A_t$ represents the assortment offered at round $t$, and $c_t$ is the item selected by the customer from $\A_t$. Formally, we let
$\mathbf{D}=\langle (c_1,\A_1), (c_2,\A_2),\dots , (c_T,\A_T) \rangle$; where for $t=1,2,\dots ,T$ we have 
$c_t=\choice_\tau(\A_t)$, $\tau\sim \cD$. While in traditional learning algorithms $\mathbf{D}$ is given as input. Here, we focus of \emph{active} learning and collect choice data by presenting assortments to the customers and recording observed  choices.

Our active learning algorithm for the estimation  of the center  is \BuCC. It takes as input the set of products $N$ and assortment size $\ell$ and offers a sequence of  assortments $\A_1,\A_2,\dots \A_T$ to the customers, recoding corresponding choices $c_t$ for $t=1,2,\dots, T$. A Pseudocode of \BuCC\ is presented in \cref{alg:bucchoi-ell}.
\begin{figure}[h]
\centering
\begin{minipage}[t]{0.41\textwidth}
\footnotesize 
\begin{algorithm}[H]
\caption{\FindTop}
\label{alg:find-top-el}
\begin{algorithmic}[1]
\STATE \textbf{Input:} Assortment $\A$, a sequence $(c_1,c_2,\dots, c_m)$ where $c_i=\choice_{\tau}(\A)$ for $\tau$ sampled from $\D$.
\STATE \textbf{Output:}  $\choice_{\tau^*}(\A)$ if $\tau^* \cap \A^{\Null} \neq \emptyset$. 
\STATE $X \gets \textbf{0}_{\A^{\Null} \times \A^{\Null}}$
\FOR{$i=1:m$}
\FOR{$a\in \A^{\Null}\setminus\{c_i\}$}
    \STATE $X_{c_i a}= X_{c_i a}+1$
    \STATE $X_{a c_i}= X_{a c_i}-1$
\ENDFOR
\ENDFOR
\STATE $Y \gets X / m$
\IF{$\exists a:Y_{aa'}> \frac{1+\left|\A\right|}{2}\forall a'\in \A^{\Null}\setminus\{a\}$}
    \STATE \textbf{Return} $a$
\ELSE
    \STATE \textbf{Return} None
\ENDIF
\end{algorithmic}
\end{algorithm}
\end{minipage}
\hfill
\begin{minipage}[t]{0.58\textwidth}
\footnotesize 
\begin{algorithm}[H]
\caption{\BuCC}
\label{alg:bucchoi-ell}
\begin{algorithmic}[1]
\STATE \textbf{Input:} $N$: set of products, $r$: assortment size,  $m$: number of samples, choice oracle $\choice_\tau;~ \tau\sim$ \topkgmm~$\D$. 
\STATE \textbf{Output:} $k$ size of center, $\tau^*$ center of $\D$
\STATE $T=\emptyset$, $B=\emptyset$, $U=N$
\REPEAT
\STATE $\A=$  assortment of size $r$ from $U$\footnote{If $\left|U\right|< r$ use some elements from $B$} 
\STATE $S=\emptyset$ {\color{blue}\footnotesize \# collect choice data $S$ by showing  $\A$ repeatedly}
\FOR{$j=1:m$}
\STATE $S=S\cup\choice_\tau (\A)$
\ENDFOR
\STATE $a=\FindTop(\A,S)$ 
\IF{$a\neq \text{None}$}
\STATE $T=T\cup\{a\}$, $U=U\setminus \{a\}$
\ELSE
\STATE $B=B\cup \A$, $U=U\setminus \A$
\ENDIF
\UNTIL{$U=\emptyset$}
\STATE $k=\left|U\right|$
\STATE $\tau^*=$ \textsc{SortCntr}$(U,r,m)$ {\color{blue}\footnotesize\# (\cref{alg:sort})}
\STATE \textbf{Return } $k,\tau^*$
\end{algorithmic}
\end{algorithm}
\end{minipage}
\end{figure}

 Assume \topkgmm\ distribution $\D$ with center $\tau^*$, and parameters $\beta$, $p$, $\vec{w}$. 
Let $w_{\min}\doteq \min_{i\in k} w_i$, and 
$n=\left|N\right|$. The following theorem shows the sample complexity of \BuCC:





\begin{theorem}\label{thm:learn2}
Assume that $\beta\geq \log 3/w_{\min}$ and $\Null\notin \tau^*$. By only receiving $N$ and $r$ as input and being able to collect choice samples $\mathbf{D}$ by selecting assortments, with probability at least $1-o(1)$, we are able to learn $\tau^*$ and $k$ from $\mathbf{D}$ using only $\Theta(r^2 \log n)$ choice samples. 
\end{theorem}


The main building block of \BuCC\ is a procedure  \FindTop\ which given an assortment $\A$ and a set of choice samples $\mathbf{D}$, outputs $i\in \A^{\Null}$ such that $i$ has the highest rank w.r.t. $\tau^*$ among elements of $\A^{\Null}$. If all elements of $\A^{\Null}$ are incomparable in $\tau^*$, \FindTop\ returns None. We note that \FindTop\ in \emph{not} an active learning algorithm and receives $\mathbf{D}$ as input. We also assume that the choice data were collected by presenting a single arbitrary assortment $\A$.

\vspace{-0.2cm}
\paragraph{\FindTop}
A pseudocode for \FindTop\ is presented in \cref{alg:find-top-el}. The main idea is to  maintain for each $i,j\in \A^{\Null}$, a variable $X_{ij}$. For any choice sample $c_t$ we increment $X_{ij}$ if $i$ is chosen over $j$, i.e., $i=c_t$ and $j\in \A\setminus\{c_t\}$, and we decrement it otherwise.  Taking $Y_{ij}=X_{ij}/m$ for all $i,j\in \A^{\Null}$, it is not difficult to see that:
\[\mathbb{E}\left[Y_{ij}\right]= \choice_\tau(i,\A)-\choice_\tau(j,\A),\quad \tau\sim \D\enspace.\]
Based on this observation and by calculating a lower bound on $\choice_\tau(i,\A)-\choice_\tau(j,\A)$ when $i$ is $\A^\Null$'s top element w.r.t. $\tau^*$ and $j\in \A^{\Null}\setminus \{i\}$ we are able to show that the top element will be found by \FindTop\ if the discrepancy parameter $\beta$ is large enough and we have enough samples.  Formally we show the following lemma whose proof is presented in full details in  \cref{app:learncenter}:

\begin{lemma}\label{lem:findtopanalysis}
Assume that $\beta\geq \log(3)/w_{\min}$ and let $r=\vert\A\vert$ and $\zeta\geq 1$ arbitrary constant. If $\A$ appears at least $\Theta( \zeta (r+1)^2\log n)$ times among the displayed assortments,
    with probability at least $1-o(n^{-\zeta})$ we have: 
    if $\A^{\Null}\cap \tau^*\neq \emptyset$, \FindTop\ will return   $i\in \A^{\Null}$ such that $i\succ_{\tau^*} j$ for any $j\in \A^{\Null}\setminus \{i\}$, otherwise, it returns None, and we can conclude that $\A^{\Null}\cap \tau^*=\emptyset$~.
\end{lemma}

\paragraph{\BuCC} 

\BuCC\ first identifies all elements in $N$ which are ranked above 
$\Null$ and are in $\tau^*$.
To this end, we maintain three sets: $T\subseteq \tau^*$ and $B\subseteq \bar{\tau^*}$, $U$ unknown whether they are in $\tau^*$ or $\bar{\tau^*}$. Assortments of size $r$ are selected repeatedly and after calling \FindTop\ we either find the top element in the assortment -- which has to be in $\tau^*$ or we find out that none of the  elements in the assortment are in $\tau^*$. 
Note that the number of times that Repeat loop iterates is bounded by $k+n/r$.
Finally we call \cref{alg:sort} to find the rank of items in $\tau^*$.
 \Cref{thm:learn2} will be concluded from \cref{lem:findtopanalysis}
and using a union bound on all \FindTop\ calls. We remark that if $\Null \in \tau^*$, \BuCC\ will be able to return the top prefix of  $\tau^*$ constituting of the elements ranked above $\Null$ (see \cref{app:learncenter}).





\vspace{-0.2cm}
\section{Experiments}\label{sec:exp}
\vspace{-0.2cm}
In this section, we present our experimental analysis, designed to achieve two main objectives: (1) to compare the predictive power of the top-$k$ Mallows model (\topkgmm) with that of the multinomial logit model (MNL), and (2) to evaluate the accuracy and computational complexity of our methods, namely \PRIM\ sampling algorithm, the \DeepCheep\ choice probability computation, and the two learning algorithms, \FindTop\ and \BuCC.  The code and log files are available publicly\footnote{Link to the code \url{https://github.com/ShahrzadGit/topkmallows-choices}}. Results are generated by running the code on a MacBook Pro M1 Max, 32GM RAM. 

\vspace{-0.2cm}
\paragraph{Predictive power of top-$k$ MM compared to MNL: experiments on real-world data}
We used ``Sushi Preference Data Set''\citep{sushidataset2005} which contains preference of customers over a set of 100 different sushi types \footnote{Link to of Sushi Preference Data Set \url{https://www.kamishima.net/sushi/}. }. The data-set includes 5K  preferences in the form of top-$10$ lists.

\begin{figure}[h]
    \centering

    \begin{subfigure}[b]{0.4\linewidth}
        \centering
    \includegraphics[width=\linewidth]{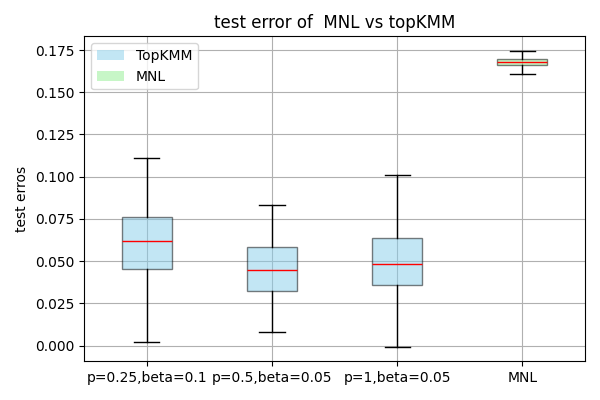}
    \caption{\scriptsize{No clustering}}
        \label{fig:test:nocl}
    \end{subfigure}\quad\quad
    \begin{subfigure}[b]{0.4\linewidth}
        \centering
        \includegraphics[width=\linewidth]{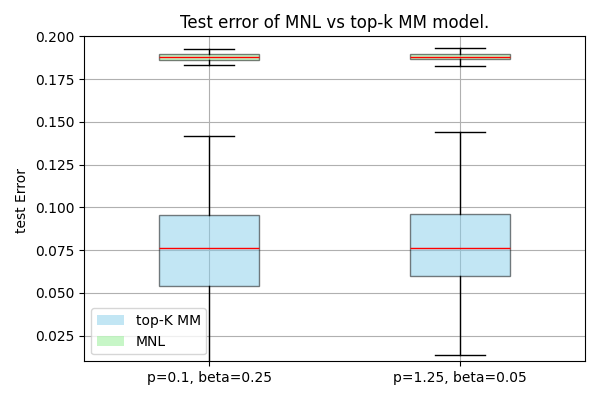}
        \caption{\scriptsize{ Training-set partitioned into two clusters}}
        \label{fig:test:twocl}
    \end{subfigure}

    \caption{\footnotesize{Test error of the top-$k$ Mallows model compared to MNL. Parameter $\beta$ and $p$ have been selected to derive highest accuracy. \Cref{table:allparams_onecluster,table:testerror-allbetas} show the test error for all choices of $p$ and $\beta$.}}
    \label{fig:gen_error}
\end{figure}

\emph{Set-Up.}
We begin by randomly splitting the 5K top-$10$
 preference data into a training set ($80\%$) and a test set ($20\%$). We apply \BuCC\ using assortments of size one or two (\cref{alg:bucchoi}) to the training set using various values of $p$ and 
$\beta$ to learn the center of the distribution. With the learned parameters, we use \DeepCheep\ to compute the corresponding choice probabilities.\\
For evaluation, we use empirical choice probabilities on the test set by repeatedly sampling random assortments and recording corresponding choices. These empirical estimates are then compared to the predictions from \DeepCheep\ to assess out-of-sample accuracy; the errors are reported in \Cref{table:allparams_onecluster}. Based on these results, we identify the values of 
$p$ and $\beta$ that yield the lowest test error. \Cref{fig:test:nocl} shows the prediction error of the \topkgmm\ compared to MNL  after tuning.

Considering multiple customer types, we cluster 
the training data (into 2–5 groups) using the Kendall's Tau distance 
$\K^{p}$
 , varying 
$p$. Clusters with positive silhouette scores are retained, and choice probabilities are computed within each cluster using both \topkgmm\ and MNL. Final predictions are weighted averages based on cluster sizes.  \Cref{fig:test:twocl} shows the results for the two-clusters.



\vspace{-0.2cm}

\emph{Experimental Findings.} Our findings show high
 accuracy of out-of-sample choice probability prediction of \topkgmm\ compared to the accuracy obtained from MNL. These results are consistent with findings of \cite{desir2021mallows} who observe the same for classic Mallows model on permutations.

\vspace{-0.2cm}
\paragraph{Accuracy and complexity of algorithms: experiments on synthetic Data}
We use synthetic data to evaluate the accuracy and complexity of our algorithms, as it provides access to ground-truth choice probabilities and distribution centers—information unavailable in real-world datasets. This enables controlled analysis of sample complexity trade-offs with respect to key parameters: $n$ (number of products), $k$ (size of top-$k$), $r$ (assortment size), $\beta$ (decay parameter) and $p$ (Kendall's Tau parameter).  

\begin{figure}[h]
    \centering
     \begin{subfigure}[b]{0.42\linewidth}
        \centering
  \includegraphics[width=\linewidth]{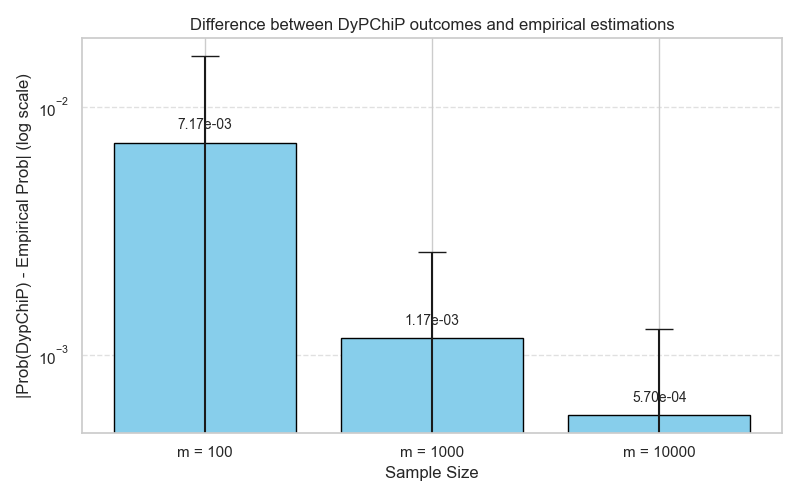}
        \caption{\scriptsize{Accuracy of \DeepCheep\ for  $n=200, k=6, r=4, p=0.5, w=2\vec{1}$.  }}
        \label{fig:accdypchip}
    \end{subfigure}\quad
   \begin{subfigure}[b]{0.39\linewidth}
        \centering
      \includegraphics[width=\linewidth]{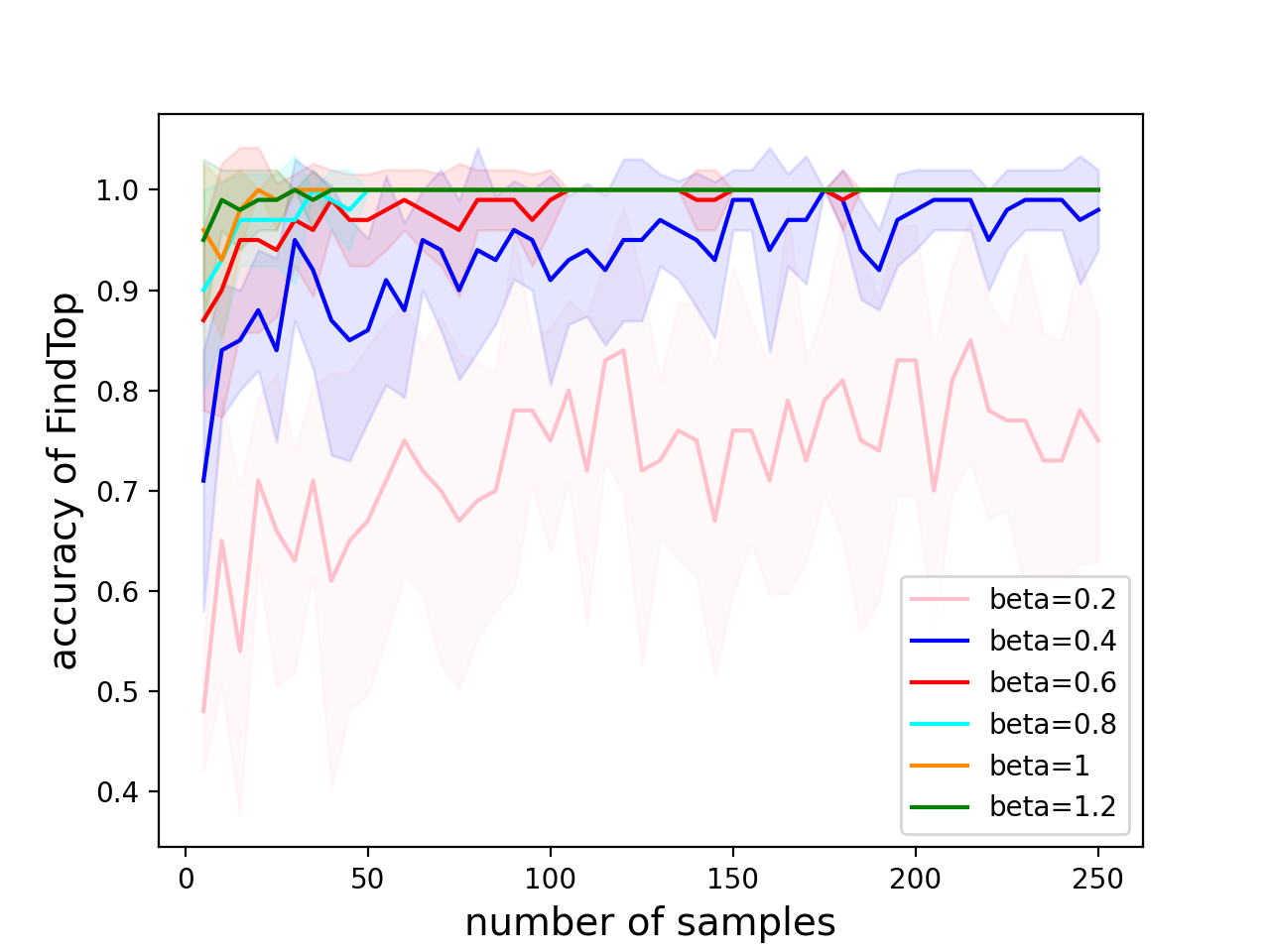}
        \caption{\scriptsize{ Sample complexity of \FindTop, $n=900, k=10, r=5,p=1, w=2\vec{1}$.   }}
        \label{fig:learnfindtop}
    \end{subfigure}
    \begin{subfigure}[b]{0.39\linewidth}
        \centering
  \includegraphics[width=\linewidth]{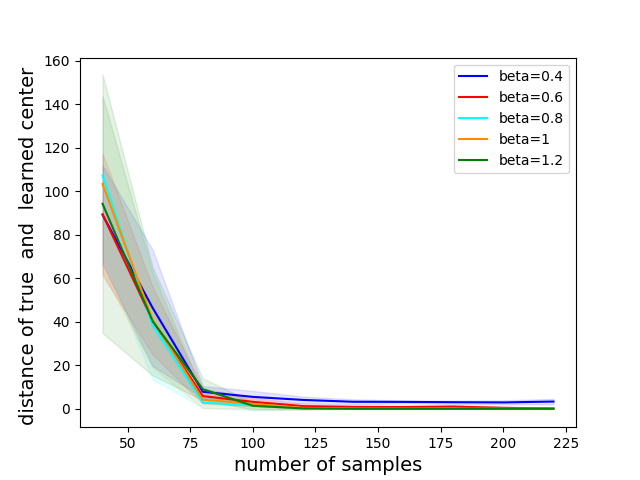}
        \caption{\scriptsize{Sample complexity of \BuCC\ for  $n=500, k=10, p=0.5, w=2\vec{1}$.  }}
        \label{fig:learnbucc}
    \end{subfigure}\quad\quad
       \begin{subfigure}[b]{0.39\linewidth}
        \centering
  \includegraphics[width=\linewidth]{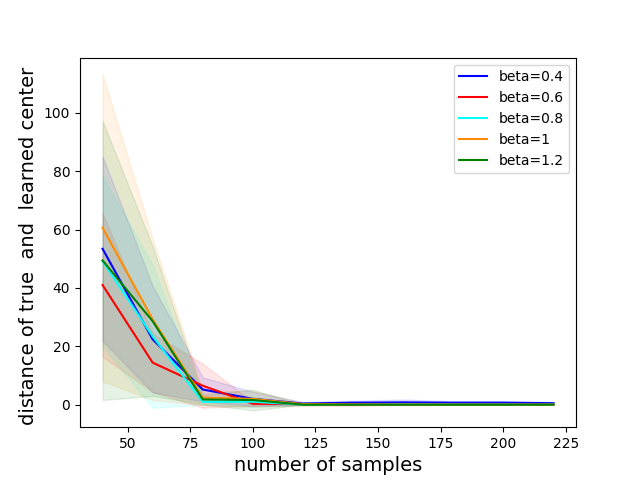}
        \caption{\scriptsize{Sample complexity of \BuCC\ for $n=300, k=8, p=2,  w=32222111$.  }}
    \end{subfigure}

    \caption{\footnotesize{performance of algorithms \DeepCheep, \FindTop\ and \BuCC\ on synthetic data. }}
     \label{fig:synth}
\end{figure}

\emph{Accuracy and time complexity of \PRIM\ and \DeepCheep.} We evaluate algorithm accuracy by generating samples from a \topkgmm\ using \PRIM\ and comparing empirical choice frequencies over random assortments to the probabilities predicted by \DeepCheep. This is repeated across 20 assortments, with the mean and standard deviation of results shown in \cref{fig:accdypchip}. The run-times of \DeepCheep\ and \PRIM\ are reported in \Cref{sec:app:runtimes}.

\emph{Accuracy and sample complexity of learning algorithms of \BuCC\ and \FindTop.} We evaluate our two learning algorithms by generating $m$ samples from \topkgmm\ distribution using \PRIM\ and running \FindTop\ and \BuCC\ to learn the top element or distribution center. We assess the convergence of these methods by comparing the learned and true values. When learning the center in \BuCC, we use the Kendall's Tau distance $\K^p$ of learned and  true center. In \FindTop\ we check whether the learned top element is the same as the true top element and directly calculate accuracy based on the frequency of matching values. Each experiments is repeated 10 times across a range of model parameters and average and standard deviation are obtained; see \cref{fig:synth}, and \cref{sec:app:learningalgs}.

\vspace{-0.4cm}
\vspace{-0.2cm}
\emph{Experimental Findings.} Our experiments in this section support the theoretical results on the accuracy and complexity of our algorithms. Notably, we observe that our methods achieve high accuracy with a relatively small number of samples—often logarithmic in the number of items 
$n$.\\
For \DeepCheep, we find that time complexity increases rapidly with 
$k$, as expected due to its exponential dependence on this parameter (See \cref{fig:dypchipruntimes}). In contrast, the runtime shows minimal sensitivity to the number of products 
$n$. For \PRIM, the exponential dependence on 
$k$ is less restrictive since it primarily affects the preprocessing step; once this step is completed, generating a large number of samples remains efficient with low amortized cost  (See \cref{tab:prim-runtimes}).\\
In the learning algorithms \BuCC\ and \FindTop, we observe that the sample complexity increases as 
$\beta$ decreases. This is expected, as smaller values of 
$\beta$ cause the distribution to approach uniformity, reducing the concentration of samples around the center and making learning more challenging.

\vspace{-0.4 cm}
\section{Conclusion}
\vspace{-0.2 cm}
In conclusion, the generalized Mallows model for top-$k$ lists provides a more realistic framework for understanding user preferences, particularly when users care only about a limited set of options. 
Our work centers on applying \topkgmm\ to choice modeling and developing several key algorithms. An important open direction is learning the model parameters for mixture versions of these models.

\vspace{-0.3 cm}
\section{Acknowledgment}
\vspace{-0.2 cm}
We are thankful to Anonymous NeurIPS reviewers for helping us strengthen the presentation of our paper. Shahrzad Haddadan is supported by Rutgers Business School's Dean’s Research Seed Fund. 

\bibliography{main.bib}
\bibliographystyle{icml2025}


\newpage
\newpage
\section*{NeurIPS Paper Checklist}

\begin{enumerate}

\item {\bf Claims}
    \item[] Question: Do the main claims made in the abstract and introduction accurately reflect the paper's contributions and scope?
    \item[] Answer: \answerYes{} 
    \item[] Justification: All claims in the introduction and abstract are later clearly stated in the main body of the paper  as theorems with a clear statement of all theoretical assumptions. 
    \item[] Guidelines:
    \begin{itemize}
        \item The answer NA means that the abstract and introduction do not include the claims made in the paper.
        \item The abstract and/or introduction should clearly state the claims made, including the contributions made in the paper and important assumptions and limitations. A No or NA answer to this question will not be perceived well by the reviewers. 
        \item The claims made should match theoretical and experimental results, and reflect how much the results can be expected to generalize to other settings. 
        \item It is fine to include aspirational goals as motivation as long as it is clear that these goals are not attained by the paper. 
    \end{itemize}

\item {\bf Limitations}
    \item[] Question: Does the paper discuss the limitations of the work performed by the authors?
    \item[] Answer: \answerYes{}
    \item[] Justification: 
The runtime of some algorithms are large and this is explained in the experiments sections.
    \item[] Guidelines:
    \begin{itemize}
        \item The answer NA means that the paper has no limitation while the answer No means that the paper has limitations, but those are not discussed in the paper. 
        \item The authors are encouraged to create a separate "Limitations" section in their paper.
        \item The paper should point out any strong assumptions and how robust the results are to violations of these assumptions (e.g., independence assumptions, noiseless settings, model well-specification, asymptotic approximations only holding locally). The authors should reflect on how these assumptions might be violated in practice and what the implications would be.
        \item The authors should reflect on the scope of the claims made, e.g., if the approach was only tested on a few datasets or with a few runs. In general, empirical results often depend on implicit assumptions, which should be articulated.
        \item The authors should reflect on the factors that influence the performance of the approach. For example, a facial recognition algorithm may perform poorly when image resolution is low or images are taken in low lighting. Or a speech-to-text system might not be used reliably to provide closed captions for online lectures because it fails to handle technical jargon.
        \item The authors should discuss the computational efficiency of the proposed algorithms and how they scale with dataset size.
        \item If applicable, the authors should discuss possible limitations of their approach to address problems of privacy and fairness.
        \item While the authors might fear that complete honesty about limitations might be used by reviewers as grounds for rejection, a worse outcome might be that reviewers discover limitations that aren't acknowledged in the paper. The authors should use their best judgment and recognize that individual actions in favor of transparency play an important role in developing norms that preserve the integrity of the community. Reviewers will be specifically instructed to not penalize honesty concerning limitations.
    \end{itemize}

\item {\bf Theory assumptions and proofs}
    \item[] Question: For each theoretical result, does the paper provide the full set of assumptions and a complete (and correct) proof?
    \item[] Answer: \answerYes{} 
    \item[] Justification: The assumptions are clearly stated and all the proofs are presented in the appendix. 
    \item[] Guidelines:
    \begin{itemize}
        \item The answer NA means that the paper does not include theoretical results. 
        \item All the theorems, formulas, and proofs in the paper should be numbered and cross-referenced.
        \item All assumptions should be clearly stated or referenced in the statement of any theorems.
        \item The proofs can either appear in the main paper or the supplemental material, but if they appear in the supplemental material, the authors are encouraged to provide a short proof sketch to provide intuition. 
        \item Inversely, any informal proof provided in the core of the paper should be complemented by formal proofs provided in appendix or supplemental material.
        \item Theorems and Lemmas that the proof relies upon should be properly referenced. 
    \end{itemize}

    \item {\bf Experimental result reproducibility}
    \item[] Question: Does the paper fully disclose all the information needed to reproduce the main experimental results of the paper to the extent that it affects the main claims and/or conclusions of the paper (regardless of whether the code and data are provided or not)?
    \item[] Answer: \answerYes{} 
    \item[] Justification: The details of set-up, evaluation metrics and the data-set are explained in detail in the appedix of the paper. 
    \item[] Guidelines:
    \begin{itemize}
        \item The answer NA means that the paper does not include experiments.
        \item If the paper includes experiments, a No answer to this question will not be perceived well by the reviewers: Making the paper reproducible is important, regardless of whether the code and data are provided or not.
        \item If the contribution is a dataset and/or model, the authors should describe the steps taken to make their results reproducible or verifiable. 
        \item Depending on the contribution, reproducibility can be accomplished in various ways. For example, if the contribution is a novel architecture, describing the architecture fully might suffice, or if the contribution is a specific model and empirical evaluation, it may be necessary to either make it possible for others to replicate the model with the same dataset, or provide access to the model. In general. releasing code and data is often one good way to accomplish this, but reproducibility can also be provided via detailed instructions for how to replicate the results, access to a hosted model (e.g., in the case of a large language model), releasing of a model checkpoint, or other means that are appropriate to the research performed.
        \item While NeurIPS does not require releasing code, the conference does require all submissions to provide some reasonable avenue for reproducibility, which may depend on the nature of the contribution. For example
        \begin{enumerate}
            \item If the contribution is primarily a new algorithm, the paper should make it clear how to reproduce that algorithm.
            \item If the contribution is primarily a new model architecture, the paper should describe the architecture clearly and fully.
            \item If the contribution is a new model (e.g., a large language model), then there should either be a way to access this model for reproducing the results or a way to reproduce the model (e.g., with an open-source dataset or instructions for how to construct the dataset).
            \item We recognize that reproducibility may be tricky in some cases, in which case authors are welcome to describe the particular way they provide for reproducibility. In the case of closed-source models, it may be that access to the model is limited in some way (e.g., to registered users), but it should be possible for other researchers to have some path to reproducing or verifying the results.
        \end{enumerate}
    \end{itemize}

\item {\bf Open access to data and code}
    \item[] Question: Does the paper provide open access to the data and code, with sufficient instructions to faithfully reproduce the main experimental results, as described in supplemental material?
    \item[] Answer: \answerYes{} 
    \item[] Justification: We have used publicly available datasets and cited them in our paper. The code is attached to this submission. 
    \item[] Guidelines:
    \begin{itemize}
        \item The answer NA means that paper does not include experiments requiring code.
        \item Please see the NeurIPS code and data submission guidelines (\url{https://nips.cc/public/guides/CodeSubmissionPolicy}) for more details.
        \item While we encourage the release of code and data, we understand that this might not be possible, so “No” is an acceptable answer. Papers cannot be rejected simply for not including code, unless this is central to the contribution (e.g., for a new open-source benchmark).
        \item The instructions should contain the exact command and environment needed to run to reproduce the results. See the NeurIPS code and data submission guidelines (\url{https://nips.cc/public/guides/CodeSubmissionPolicy}) for more details.
        \item The authors should provide instructions on data access and preparation, including how to access the raw data, preprocessed data, intermediate data, and generated data, etc.
        \item The authors should provide scripts to reproduce all experimental results for the new proposed method and baselines. If only a subset of experiments are reproducible, they should state which ones are omitted from the script and why.
        \item At submission time, to preserve anonymity, the authors should release anonymized versions (if applicable).
        \item Providing as much information as possible in supplemental material (appended to the paper) is recommended, but including URLs to data and code is permitted.
    \end{itemize}

\item {\bf Experimental setting/details}
    \item[] Question: Does the paper specify all the training and test details (e.g., data splits, hyperparameters, how they were chosen, type of optimizer, etc.) necessary to understand the results?
    \item[] Answer: \answerYes{} 
    \item[] Justification: All hyperparameter settings are explained in detail in the appendix.  
    \item[] Guidelines:
    \begin{itemize}
        \item The answer NA means that the paper does not include experiments.
        \item The experimental setting should be presented in the core of the paper to a level of detail that is necessary to appreciate the results and make sense of them.
        \item The full details can be provided either with the code, in appendix, or as supplemental material.
    \end{itemize}

\item {\bf Experiment statistical significance}
    \item[] Question: Does the paper report error bars suitably and correctly defined or other appropriate information about the statistical significance of the experiments?
    \item[] Answer: \answerNo{}
    \item[] Justification: We have ran all experiments 5 times and average results are reported but we don't have statistical significance analysis. We don't think it is applicable to our problem. 
    \item[] Guidelines:
    \begin{itemize}
        \item The answer NA means that the paper does not include experiments.
        \item The authors should answer "Yes" if the results are accompanied by error bars, confidence intervals, or statistical significance tests, at least for the experiments that support the main claims of the paper.
        \item The factors of variability that the error bars are capturing should be clearly stated (for example, train/test split, initialization, random drawing of some parameter, or overall run with given experimental conditions).
        \item The method for calculating the error bars should be explained (closed form formula, call to a library function, bootstrap, etc.)
        \item The assumptions made should be given (e.g., Normally distributed errors).
        \item It should be clear whether the error bar is the standard deviation or the standard error of the mean.
        \item It is OK to report 1-sigma error bars, but one should state it. The authors should preferably report a 2-sigma error bar than state that they have a 96\% CI, if the hypothesis of Normality of errors is not verified.
        \item For asymmetric distributions, the authors should be careful not to show in tables or figures symmetric error bars that would yield results that are out of range (e.g. negative error rates).
        \item If error bars are reported in tables or plots, The authors should explain in the text how they were calculated and reference the corresponding figures or tables in the text.
    \end{itemize}

\item {\bf Experiments compute resources}
    \item[] Question: For each experiment, does the paper provide sufficient information on the computer resources (type of compute workers, memory, time of execution) needed to reproduce the experiments?
    \item[] Answer: \answerYes{} 
    \item[] Justification: They are reported at the beginning of the experiments section. 
    \item[] Guidelines:
    \begin{itemize}
        \item The answer NA means that the paper does not include experiments.
        \item The paper should indicate the type of compute workers CPU or GPU, internal cluster, or cloud provider, including relevant memory and storage.
        \item The paper should provide the amount of compute required for each of the individual experimental runs as well as estimate the total compute. 
        \item The paper should disclose whether the full research project required more compute than the experiments reported in the paper (e.g., preliminary or failed experiments that didn't make it into the paper). 
    \end{itemize}
    
\item {\bf Code of ethics}
    \item[] Question: Does the research conducted in the paper conform, in every respect, with the NeurIPS Code of Ethics \url{https://neurips.cc/public/EthicsGuidelines}?
    \item[] Answer: \answerYes{} 
    \item[] Justification: It does. 
    \item[] Guidelines:
    \begin{itemize}
        \item The answer NA means that the authors have not reviewed the NeurIPS Code of Ethics.
        \item If the authors answer No, they should explain the special circumstances that require a deviation from the Code of Ethics.
        \item The authors should make sure to preserve anonymity (e.g., if there is a special consideration due to laws or regulations in their jurisdiction).
    \end{itemize}

\item {\bf Broader impacts}
    \item[] Question: Does the paper discuss both potential positive societal impacts and negative societal impacts of the work performed?
    \item[] Answer: \answerNA{} 
    \item[] Justification: We are not aware of any negative societal impact. 
    \item[] Guidelines:
    \begin{itemize}
        \item The answer NA means that there is no societal impact of the work performed.
        \item If the authors answer NA or No, they should explain why their work has no societal impact or why the paper does not address societal impact.
        \item Examples of negative societal impacts include potential malicious or unintended uses (e.g., disinformation, generating fake profiles, surveillance), fairness considerations (e.g., deployment of technologies that could make decisions that unfairly impact specific groups), privacy considerations, and security considerations.
        \item The conference expects that many papers will be foundational research and not tied to particular applications, let alone deployments. However, if there is a direct path to any negative applications, the authors should point it out. For example, it is legitimate to point out that an improvement in the quality of generative models could be used to generate deepfakes for disinformation. On the other hand, it is not needed to point out that a generic algorithm for optimizing neural networks could enable people to train models that generate Deepfakes faster.
        \item The authors should consider possible harms that could arise when the technology is being used as intended and functioning correctly, harms that could arise when the technology is being used as intended but gives incorrect results, and harms following from (intentional or unintentional) misuse of the technology.
        \item If there are negative societal impacts, the authors could also discuss possible mitigation strategies (e.g., gated release of models, providing defenses in addition to attacks, mechanisms for monitoring misuse, mechanisms to monitor how a system learns from feedback over time, improving the efficiency and accessibility of ML).
    \end{itemize}
    
\item {\bf Safeguards}
    \item[] Question: Does the paper describe safeguards that have been put in place for responsible release of data or models that have a high risk for misuse (e.g., pretrained language models, image generators, or scraped datasets)?
    \item[] Answer: \answerNA{} 
    \item[] Justification: We belive that our work does not have such risks.
    \item[] Guidelines:
    \begin{itemize}
        \item The answer NA means that the paper poses no such risks.
        \item Released models that have a high risk for misuse or dual-use should be released with necessary safeguards to allow for controlled use of the model, for example by requiring that users adhere to usage guidelines or restrictions to access the model or implementing safety filters. 
        \item Datasets that have been scraped from the Internet could pose safety risks. The authors should describe how they avoided releasing unsafe images.
        \item We recognize that providing effective safeguards is challenging, and many papers do not require this, but we encourage authors to take this into account and make a best faith effort.
    \end{itemize}

\item {\bf Licenses for existing assets}
    \item[] Question: Are the creators or original owners of assets (e.g., code, data, models), used in the paper, properly credited and are the license and terms of use explicitly mentioned and properly respected?
    \item[] Answer: \answerNA{} 
    \item[] Justification: We have only used publicly available datasets which are properly cited. 
    \item[] Guidelines:
    \begin{itemize}
        \item The answer NA means that the paper does not use existing assets.
        \item The authors should cite the original paper that produced the code package or dataset.
        \item The authors should state which version of the asset is used and, if possible, include a URL.
        \item The name of the license (e.g., CC-BY 4.0) should be included for each asset.
        \item For scraped data from a particular source (e.g., website), the copyright and terms of service of that source should be provided.
        \item If assets are released, the license, copyright information, and terms of use in the package should be provided. For popular datasets, \url{paperswithcode.com/datasets} has curated licenses for some datasets. Their licensing guide can help determine the license of a dataset.
        \item For existing datasets that are re-packaged, both the original license and the license of the derived asset (if it has changed) should be provided.
        \item If this information is not available online, the authors are encouraged to reach out to the asset's creators.
    \end{itemize}

\item {\bf New assets}
    \item[] Question: Are new assets introduced in the paper well documented and is the documentation provided alongside the assets?
    \item[] Answer: \answerNA{} 
    \item[] Justification: We don't release new datasets. 
    \item[] Guidelines:
    \begin{itemize}
        \item The answer NA means that the paper does not release new assets.
        \item Researchers should communicate the details of the dataset/code/model as part of their submissions via structured templates. This includes details about training, license, limitations, etc. 
        \item The paper should discuss whether and how consent was obtained from people whose asset is used.
        \item At submission time, remember to anonymize your assets (if applicable). You can either create an anonymized URL or include an anonymized zip file.
    \end{itemize}

\item {\bf Crowdsourcing and research with human subjects}
    \item[] Question: For crowdsourcing experiments and research with human subjects, does the paper include the full text of instructions given to participants and screenshots, if applicable, as well as details about compensation (if any)? 
    \item[] Answer: \answerNA{} 
    \item[] Justification: human subjects
    \item[] Guidelines:
    \begin{itemize}
        \item The answer NA means that the paper does not involve crowdsourcing nor research with human subjects.
        \item Including this information in the supplemental material is fine, but if the main contribution of the paper involves human subjects, then as much detail as possible should be included in the main paper. 
        \item According to the NeurIPS Code of Ethics, workers involved in data collection, curation, or other labor should be paid at least the minimum wage in the country of the data collector. 
    \end{itemize}

\item {\bf Institutional review board (IRB) approvals or equivalent for research with human subjects}
    \item[] Question: Does the paper describe potential risks incurred by study participants, whether such risks were disclosed to the subjects, and whether Institutional Review Board (IRB) approvals (or an equivalent approval/review based on the requirements of your country or institution) were obtained?
    \item[] Answer: \answerNA{} 
    \item[] Justification: Our paper does not involve
    human subjects
    \item[] Guidelines:
    \begin{itemize}
        \item The answer NA means that the paper does not involve crowdsourcing nor research with human subjects.
        \item Depending on the country in which research is conducted, IRB approval (or equivalent) may be required for any human subjects research. If you obtained IRB approval, you should clearly state this in the paper. 
        \item We recognize that the procedures for this may vary significantly between institutions and locations, and we expect authors to adhere to the NeurIPS Code of Ethics and the guidelines for their institution. 
        \item For initial submissions, do not include any information that would break anonymity (if applicable), such as the institution conducting the review.
    \end{itemize}

\item {\bf Declaration of LLM usage}
    \item[] Question: Does the paper describe the usage of LLMs if it is an important, original, or non-standard component of the core methods in this research? Note that if the LLM is used only for writing, editing, or formatting purposes and does not impact the core methodology, scientific rigorousness, or originality of the research, declaration is not required.
    \item[] Answer: \answerNA{} 
    \item[] Justification: 
    \item[] Guidelines:
    \begin{itemize}
        \item The answer NA means that the core method development in this research does not involve LLMs as any important, original, or non-standard components.
        \item Please refer to our LLM policy (\url{https://neurips.cc/Conferences/2025/LLM}) for what should or should not be described.
    \end{itemize}

\end{enumerate}

\appendix

\newpage 
\section{Missing proofs and details}

\subsection{Missing details from 
\Cref{sec:samplealg}}\label{app:profile_proofs}
In this section, we prove \Cref{thm:sampling}. To this end, we first establish the correctness of \Cref{lem:profile_prob} that proves profile $S$ can be sampled according to $\probs{S}{\D}$. 
\begin{replemma}[Restatement of \Cref{lem:profile_prob}]
Given a \topkgmm\ distribution $\D$ with  parameters $\beta$, $p$ and $\bm{w}$, any profile $S = \{s_1, s_2, \dots, s_\ell\} \subseteq [k]$ ($s_1<s_2<\cdots<s_l$) has probability $\probs{S}{\D}$ proportional to $\exp\left(-\beta f(S) \right)Z(S)$, where: 
\begin{eqnarray*}\label{eq:profilep}
    f(S) &:= &w_0 p Q(S) + \sum_{j \in [k]\setminus S} w_j (I_j(S)+p P_j(S)) \\
    Z(S) &:= &{n-k \choose k-\ell} (k-\ell)! \prod_{j=1}^{\ell} \sum_{r=0}^{k-j} e^{-\beta w_{s_j} r} 
  \end{eqnarray*}
and it can be sampled in $O(2^k k)$.
\end{replemma}
\begin{proof}
Let $\mathcal{M}$ be the normalizing factor in \Cref{eq:topkGMM}, so $\mathcal{M}=\sum_{\tau\in T_{k,N}} \probs{\tau}{\D}$. Then
\begin{align*}
    \probs{S}{\D} &= \frac{1}{\mathcal{M}} \sum_{\tau\in T(S)} \probs{\tau}{\D}  \tag{\cref{eq:profile}}\\
    &= \frac{1}{\mathcal{M}} \sum_{\tau\in T(S)} \exp \left(-\beta (w_0 p Q(\tau) + \sum_{j\in [k]} w_j \cdot(I_j(\tau)+pP_j(\tau))) \right) \tag{\cref{eq:topkGMM}}\\
    &= \frac{1}{\mathcal{M}} \sum_{\tau\in T(S)} \exp \left(-\beta (w_0 p Q(S) + \sum_{i\in[k]\setminus S} w_j \cdot (I_j(S)+pP_j(S)) + \sum_{j\in S} w_j \cdot I_j(\tau)) \right)\\
    &= \frac{1}{\mathcal{M}} \sum_{\tau\in T(S)} \exp \left(-\beta (w_0 p Q(S) + \sum_{i\in[k]\setminus S} w_j \cdot (I_j(S)+pP_j(S))) \right) 
    \cdot \exp \left(-\beta \sum_{j\in S} w_j \cdot I_j(\tau) \right)\\
    &= \frac{\exp(-\beta f(S))}{\mathcal{M}} \sum_{\tau\in T(S)} \exp \left(-\beta \sum_{j\in S} w_j \cdot I_j(\tau) \right)
\end{align*}
The remaining sum relates to the inversions of elements in $S$. Based on \Cref{def:inversiontable}, an element $s_j\in S \subseteq [k]$ makes an inversion with another element $s$ iff $s_j\succ_{\tau^*} s$ and $s\succ_\tau s_j$. Note that since $\tau\in T_{\tau^*}(S)$, the possibilities for such $s$ are $\{s_{j+1}, \cdots, s_{\ell}\}$ and the elements of $[n]\setminus [k]$ which are now in $\tau$, i.e., $\tau\cap \{k+1, \cdots,n\}$. Note that $\tau\cap \{k+1, \cdots,n\}=k-\ell$. Thus, the inversion of $j$ can be any number between $0$ to $(\ell-j)+(k-\ell)$ which is any number from $0$ to $k-j$. Furthermore, for any valid selection of these values for elements $j\in S$ w.r.t. defined ranges, then the position of these elements in $\tau$ are uniquely determined. This can be achieved by starting with a sequence of $k-\ell$ stars which would correspond to any selection of elements from $\{k+1, \cdots, n\}$, and then inserting element $s \in (s_\ell, s_{\ell -1 }, \cdots, s_1)$ iteratively based on their value of $I_s$. This shows the 1:1 correspondence between values of $I_j$s and how elements of $S$ are positioned in $\tau$. Since there are ${n-k \choose k-l} (k-l)!$ possible cases for the arrangement of the remaining elements, we get that
\begin{equation*}
    \sum_{\tau\in T(S)} \exp \left(-\beta \sum_{j\in S} w_j \cdot I_j(\tau) \right)  =  {n-k \choose k-l} (k-l)! \sum_{\substack{\text{valid choice of } I_s \\ \text{for all } s \in S}}\exp(-\beta \sum_{j\in S} w_j\cdot I_j) = Z(S).
\end{equation*}
where the last equality follows from the fact that depending on the value of $I_{s_j}$, any of the terms in the summation $1+e^{-w_{s_j}\beta}+ \dots +e^{-(k-j)w_{s_j}\beta}$ appears once. Taking product over the terms selected from these sums would exactly correspond to a valid selection of $I_s$ for all $s\in S$. Since $\mathcal{M}=\sum_{\tau\in T_{k,N}} \probs{\tau}{\D}$ and each $\probs{\tau}{\D}$ contributes to the $\probs{S}{\cD}$, the defined values form a probability distribution over different subsets of $[k]$ where $\probs{S}{\D}$ is proportional to $\exp\left(-\beta f(S)\right) Z(S)$.

It remains to show that this value can be computed in $O(2^\gamma k)$; $\gamma=\min\{k,n-k\}$. We argue that for each $S$, $Z(S)$ and $f(S)$ can be calculated in $O(k)$, the total number of profiles is bounded by $2^\gamma$, hence all the computation can be done in $O(2^\gamma k)$. For $Z(S)$, the coefficient before sum has at most $2k$ terms and there are at most $k$ terms in the sum where each term corresponds to a geometric series, so $Z(S)$ can easily be calculated in $O(k)$. $f(S)$ can be calculated in $O(k)$ using the following algorithm which clearly has $O(k)$ runtime.

\begin{algorithm}
\caption{\algSAMPLEabb}
\label{alg:sample}
\begin{algorithmic}
\STATE \textbf{Input:} {\topkgmm\ $\cD(\beta, \bm{w})$.} 
\STATE\textbf{Output:} {Sampled Top-$k$ ordering $\tau$ proportional to $\cD$ }
\STATE  $f(S)=\algPROFILEprob$
\STATE Sample $S$ proportional to $\probs{S}{\D}=f(S)\cdot Z(S)$.
\STATE \textbf{Return} $\PRIM(\beta, \bm{w}, S)$.
\end{algorithmic}\label{alg:sample}
\end{algorithm}
\begin{algorithm}
\caption{\algPRIMabb}
\label{alg:prim}
\begin{algorithmic}
\STATE\textbf{Input:} {\topkgmm $(\beta, \bm{w})$, $S = \{s_1, s_2, \dots, s_\ell\} \subseteq [k]$ where $s_1 < s_2 <\cdots < s_{\ell}$.} 
\STATE \textbf{Output:} {Top-$k$ list $\tau \in \pss$.}
\STATE $A \gets $ ordered random $k-\ell$ elements from $[n]\setminus [k]$ 
\FOR{$s \gets s_\ell, s_{\ell-1}, \cdots, s_1$}
\STATE Insert $s$ at position $j$ in $A$ w.p.
$
\frac{\exp\left(
-\beta w_{s}j
\right)}{\sum_{x=0}^{|A|} \exp\left(-\beta w_{s}x\right)}~. 
$
\ENDFOR
\STATE \textbf{Return} $\tau = A$
\end{algorithmic}
\end{algorithm}


\begin{algorithm}
\caption{\algPROFILEprob}
\label{alg:profile}
\begin{algorithmic}
\STATE \textbf{Input:} {Profile $S = \{s_1, s_2, \dots, s_\ell\} \subseteq [k]$.} 
\STATE \textbf{Output:}{$f(S)$}
\STATE $Q(S) \gets {k - \ell \choose 2}$.
\STATE $x,y \gets 0,0$
\FOR{$j \in \{k, k-1, \cdots, 1\}$}
\STATE $I_j \gets k - \ell + x$
\STATE $P_j \gets n - 2k + \ell + y$
\IF{$j \in S$}
\STATE $x \gets x + 1$
\ELSE
\STATE $y \gets y + 1$
\ENDIF
\ENDFOR
\STATE \textbf{Return} $f(S) := w_0 p Q(S) + \sum_{j \in [k]\setminus S} w_j (I_j(S)+p P_j(S))$.
\end{algorithmic}
\end{algorithm} 
\end{proof}
Using \Cref{lem:profile_prob} and analyzing \Cref{alg:sample}, we can prove \Cref{thm:sampling}.
\begin{proof}[\textbf{Proof of \cref{thm:sampling}}]
Any $\tau\in \To{k,N}$ with $S = \tau \cap [k] $, according to \Cref{alg:prim}, is sampled with probability 
$$
\frac{1}{{n-k \choose k-l} (k-l)!} \cdot \frac{\sum_{j\in S} \exp(-\beta w_{j} I_{j})} {\sum_{j=1}^{l}\prod_{r=0}^{k-j} \exp(-\beta w_{s_j}r)} = \frac{\sum_{j\in S} \exp(-\beta w_{j} I_{j})}{Z(S)}.
$$
Since each $S$, is sampled by $\frac{\exp(-\beta f(S))Z(S)}{\sum_{S\subseteq [k]}\exp(-\beta f(S))Z(S)}$, we get that $\tau$ is sampled with 
\[
\frac{\exp(-\beta f(S))Z(S)}{\sum_{S\subseteq [k]}\exp(-\beta f(S))Z(S)}\cdot \frac{\exp(\sum_{j\in S} -\beta w_{j} I_{j})}{Z(S)} 
= \frac{\exp(-\beta (f(S) + \sum_{j\in S}- w_{j} I_{j}))}{\sum_{S\subseteq [k]}\exp(-\beta f(S))Z(S)}
\]
where the denominator as previously discussed in proof of \Cref{lem:profile_prob}, is equal to normalizing factor $\mathcal{M} = \sum_{\tau \in T_{k,N}} \probs{\tau}{\D}$. Since numerator is just restating \Cref{eq:topkGMM}, each $\tau$ is sampled w.r.t \Cref{eq:topkGMM}.

\paragraph{Time complexity} An upperbound on the number of profiles is $2^\gamma; \gamma=\min\{k,n-k\}$. Therefore $Z(S)$ and $f(S)$ can be computed in $O(2^\gamma k)$ by \Cref{lem:profile_prob}, and we can sample $S$ in $O(2^\gamma k)$. 
Next, fixing $S$, we need to execute \Cref{alg:prim}, where we first need to sample $k-\ell$ elements which an be done in $O(k \log n)$. And then inserting elements of $S$ with respect to probabilities that can be calculated in $O(k)$ for each element of $S$, so requiring $O(k^2)$ time for the recursive insertion loop. Hence overall the algorithm runs in $O(2^\gamma k + k \log n + k^2)$; $\gamma=\min\{k,n-k\}$.
\end{proof}

\subsection{Missing details from \cref{sec:choice}}\label{app:choice}
In this section, we provide the details in design of \DeepCheep.  The main remaining part is the calculation of $\pi_S(i,j,q)$ for given $\cA$, $i=1,\dots r, r+1,\dots r+\ell$,  $q=0,1,\dots \ell$, and $j=r+1,r+2,\dots r+q,k$; where $r=|\bar{\A}|$ and $\ell=|S|$. 
We remind the reader that in $\pi(i,j,q)$:
\begin{itemize}
    \item 
$i$ indicated the index of an item $a$ whose choice probability is being calculated. We use $L$ to refer to the ordered set  $\{a_1,a_2,\dots, a_r, s_\ell, s_{\ell-1},\dots, s_1\}$  a  ranking of the elements in $\bar{\cA}\cup S$ where $a_1,a_2,\dots, a_r$ is an arbitrary ranking of $\bar{\cA}$ and $s_\ell,s_{\ell-1},\dots, s_1$ is the ranking of $S$ used in \PRIM. Thus, when we say $a_i$ we mean the $i$th element in $L$.
\item $j$   indicates the position of $a_i$ if it is the winner until the $q$th iteration of the for loop in \PRIM. When $j=k$ we mean that the winner is in $\bar{\tau}$.
\item $q$ denote that so far we are considering only elements that have been sampled until the $q$th iteration of the for loop in \PRIM.
\end{itemize}
Since $\pi_S(i,j,q)$ can only take non-zero value when $a_i$ is sampled in the top-$k$ list, we either have $a_i\in \bar{\tau}$ or $a_i \in S$. We analyze each of these two possible cases separately.
We first initialize the DP table for   $q=0$ by considering $i=1,2,\dots r$. Then we consider items that are in $S$ in the reverse order of their priority (consistent with the \cref{alg:prim}). 

\begin{enumerate}
    \item 
    
     Let $\bar{\cA} = \cAN \cap \bar{\tau}$.
       Before start of the loop in \PRIM, only elements of $\bar{\cA}$ have been sampled and have nonzero probability of being selected. These elements are the first $r$ elements in our list. Thus, for $i=1,2,\dots, r$, we store these probabilities in the DP table as: 
\begin{equation}\label{eq:choiceprob:firstcond}
\begin{cases}
       \pi(i,j,0)=
    \underbrace{
    \frac{1}{n-k-j+1}}_{\substack{\text{the probability that}\\ \text{$i$ is sampled at position $j$-}}}
    \underbrace{
    \prod_{j'=1}^{j-1}(1-\frac{r}{n-k-j'})}_{\substack{\text{the probability that}\\
    \text{ no element of $\bar{\A}$ is sampled}\\
    \text{  in positions $j-1$ }}} & \text{for  } j\leq r\\
    \pi(i,k,0)= \frac{{n-k-(r+1) \choose{k-\ell}}}{{{n-k}\choose{k-\ell}}}\cdot \frac{1}{r+1} & \text{the case that $\bar{\A}\subseteq \bar{\tau}$}
\end{cases}
\end{equation}

    \item For $q=1,2,\dots, \ell$, consider the for loop in \PRIM\ and let $a_{cur}$ be the element that  is inserted in the top-$k$ list at iteration $q$ of this loop. Therefore, 
    $a_j=s_{\ell+1-q}$, and in our list it is ranked $(r+q)$th, thus, $cur=r+q$ . We distinguish between the two cases where $a_{cur}\in \cAN$ or not. 
    
\smallskip 
    \emph{Case 1.} $a_{cur}\notin \cAN$: since $a_{cur}$ is not a choice option, we have that $\pi_S(cur,j,q)=0$ for all $j\leq q$.
    For any $i<cur$, if $a_{cur}$ is inserted higher than the previous winner, the winner position will increment. If $a_{cur}$ is inserted lower than the previous winner, the winner position will not change. Therefore $\forall i< cur$, 
\[
\pi_S(i,j,q)= \pi_S(i,j,q-1)\cdot\primpos(S,>j,q)+\pi_S(i,j-1,q-1)\cdot\primpos(S,\leq j-1,q)~,
\]

where $\primpos(S,>j,q)$ is the probability that at iteration $q$ of $\PRIM$ we insert an element in a position after (lower than) $j$, and 
$\primpos(S,\leq j,q)$ is it is inserted before (higher than) $j$. Note that $\primpos(S,\leq j,q)= 1- \primpos(S,> j,q)$, and 
the  probability $\primpos(S,\leq j,q)$ can be directly calculated using \cref{eq:rimprob} by summing over all insertion probabilities for $j'=1,2,\dots,j$.

    \emph{Case 2.} $a_{cur}\in \cAN$:
    In this case, if $a_{cur}$ is positioned higher than the previous winner, it will become the new winner. Otherwise, the position of the previous winner will increment. Thus, 
    \[
    \forall i< cur,\quad \pi_S(i,j,q)=\pi_S(i,j,q-1)\cdot\primpos(S,>j,q)\enspace.
    \]
    Furthermore, 
    \[
    \pi_S(cur,j, q)= \primpos(S,j,q)\cdot \sum_{i<cur}\sum_{j'=j}^\ell \pi_S(i,j',q-1)\enspace ,
    \]
    where $\primpos(S,j,q)$  is the probability that at the $q$th iteration of \PRIM\ we insert an element in location $j$ which again can be directly calculated from  \cref{eq:rimprob}.

\end{enumerate}

\begin{proof}[\textbf{Proof of \cref{thm:choicep}}]
The above analysis forms a proof for the correctness of \DeepCheep.
 \end{proof}

\subsection{Missing proofs and details from \cref{sec:learning}}
\label{app:learncenter}

In this section we provide the missing proof for correctness and sample complexity of \BuCC\ and \FindTop\ and we present the missing details. 

Let us first present the pseudocode of 
\textsc{SortCntr} which is called in \BuCC.

\begin{algorithm}
\caption{
\textsc{SortCntr}
}\label{alg:sort}
\begin{algorithmic}
\STATE \textbf{Input:} 
\STATE choice oracle $\choice_\tau$; $\tau\sim \mathcal{D}$ where $\mathcal D$ is a topKGMM with center $\tau^*$
\STATE $U$ containing elements of $\tau^*$ as an  (unsorted) set, 
\STATE $r$ assortment size
\STATE $m$ sample size 
\STATE $B$ set of elements out of center 
\STATE \textbf{Output:} $\tau^*$ sorted
\STATE $k=\left|U\right|$
\FOR{$i=1:k$}
\STATE {\color{blue}\# find the top element in $U$ and delete it}
\STATE $T=U$
\REPEAT
 \STATE {\color{blue}\# $T$ maintains potential top elements of $U$ }
\STATE Divide $T$ to non-intersecting assortments of size $r$: $\A_1, \A_2,\dots \A_\gamma$; $\gamma=\lceil i/r\rceil$ {\color{blue}\# use elements of $B$ for $\A_\gamma$ if needed} 
\FOR{$\kappa=1:\gamma$}
\STATE $T=\emptyset$
\STATE {\color{blue} \# collect choice samples by showing $\A_\kappa$ to customers and find the top element of $\A_\kappa$}
\STATE $S=\emptyset$
\FOR{$j=1:m$}
\STATE $S=S\cup \choice_\tau(\A_\kappa)$
\ENDFOR 
\STATE $T=T\cup\{\FindTop(\A_\kappa,S)\}$
\ENDFOR
\UNTIL{$\left|T\right|=1$}
\STATE $U=U\setminus T$
\ENDFOR
\STATE \textbf{Return} $\tau^*$
\end{algorithmic}
\end{algorithm}

\paragraph{\textsc{SortCntr}} When \BuCC\ finds all the elements that are in the center, we call \textsc{SortCntr} to find the ranking of these items as a list. In the for loop the item which is positioned the $i$th will be found and deleted from $U$. In other words, we first find the top element of $U$ this is the element ranked first, after deleting it, we find the top element in the remainder which is ranked second and so on. 

In order to find the top element we use a Repeat-Until loop. In this loop   $U$ is divided to assortments of size $r$ and using \FindTop\ we find the winner in each assortment and continue by having a tournament between the winners until the winner of all (which is the top element) is found. 

Note that the repeat-until loop iterates at most $\log_r(k)$ times. We call \FindTop\ at each iteration $\lceil k/r\rceil $ times, and since we have for loop the total number of times that \FindTop\ is called is bounded by $\Theta(k^2\log_r(k))$.

We now present the proof lemmas and theorems.
As before, without loss of generality we assume that $\tau^* = 1 \succ 2 \succ \cdots k \succ \{k+1, \cdots, n\}$.

\begin{lemma}\label{lem:5.1}
Let $i<j\in \A_t$, and assume 
${\mathcal I}=\{i_1,i_2,\dots, i_\ell\}\subseteq \A_t$ is such that  $i<i_\kappa<j $ for all $i_\kappa\in \mathcal{I}$.
We have: 
\[\choice_\D(i,\A_t)\geq \choice_\D(j,\A_t)\cdot \exp\left(\beta (w_i+\sum_{\kappa=1}^\ell w_{i_\kappa})\right)\enspace.\]
\end{lemma}

Consider now a fixed $\A$ and let $a_r,a_{r-1},\dots ,a_1, \A_{0}$ be such that, for each $a_i$, there are $i$  elements in $\A^{\Null}$ which are ranked under it by $\tau^*$, i.e., $\vert\{a\in \A^{\Null}\mid a>a_i\}\vert=i$. 
As a consequence of this definition we have: $\vert\A\vert=r$ and $\A^{\Null}\cap \tau^*=\{a_r,a_{r-1},\dots , a_1\}$, and $\A_0\subseteq \bar{\tau}$. We denote any arbitrary element of $\A_0$ by $a_0$.

\begin{lemma}\label{lemma:gapbound}
For any $\kappa= r-1, r-2,\dots 1, 0$, 
we have: 
\[
\choice (a_r,\A)-
\choice(a_\kappa,\A)\geq \frac{1-\exp(- \beta w_{a_r})}{1+ r \exp(- \beta w_{a_r})}\enspace.
\]

\end{lemma}

\begin{proof}
From \cref{lem:5.1} we can immediately conclude that:
for any $\kappa= r-1, r-2,\dots 1, 0$, $\choice(a_r,\A)\cdot \exp(-\beta w_{a_r})\geq  \choice(a_\kappa,\A)$.

Let us now rearrange this inequality:

\begin{align}\label{eq:gapbound}
\choice(a_r,\A)\cdot \exp(-\beta w_{a_r})&\geq  \choice(a_\kappa,\A) \nonumber \\
\choice(a_r,\A)\cdot \exp(-\beta w_{a_r})-\choice (a_r,\A)&\geq  \choice(a_\kappa,\A)-\choice (a_r,\A) \nonumber \\
\choice(a_r,\A)[1-\exp(-\beta w_{a_r})]&\leq \choice (a_r,\A)-
\choice(a_\kappa,\A)
\end{align}

On the other hand we have that:
$   \sum_{\kappa=1}^r\choice(a_\kappa,\A)+ \sum_{a_0\in \A_0}\choice(a_0,\A)=1$.
Thus, 

\begin{align*}
   1\leq \choice(a_r,\A)+ r \choice(a_r,\A)\cdot \exp(-\beta w_{a_r}) \\
   \choice(a_r,\A)\geq 1/ (1+r \exp(-\beta w_{a_r}))~.
\end{align*}

We now substitute this lower bound in \cref{eq:gapbound} and obtain the premise. 
\end{proof}

Consider now for any for any $i,j\in \A^{\Null}$, for $t=1, 2, \cdots, T$, assume we keep displaying assortment $\A$ and consider the following random variable:

$$
X^{t}_{ij}=
\begin{cases}
+1 & \text{if } i \text{ is chosen }\\
-1 & \text{if } j \text{ is chosen }\\
0 & \text{otherwise}
\end{cases}
$$

We denote $Y_{ij}= \sum_{t=1}^T X^t_{ij}/T$, we have that: 
\[
\mathbb{E}\left[Y_{ij}\right]= \choice(i,\A)-\choice(j,\A)~.
\]

In the following lemma we let $w_{\min}$ be the minium weight assigned to the elements in $\tau^*$, i.e., $w_{\min}\doteq \min_{i\in k} w_i$~.

\begin{lemma}[Restatement of \cref{lem:findtopanalysis}]
Assume that $w_{\min}\geq \log(3)/\beta$ and let $r=\vert\A\vert$ and $\zeta\geq 1$ arbitrary constant. If $\A$ appears at least $\Theta( \zeta (r+1)^2\log n)$ times among the displayed assortments,
    with probability at least $1-o(n^{-\zeta})$ we have: 
    if $\A^{\Null}\cap \tau^*\neq \emptyset$ we are able to identify $i\in \A^{\Null}$ such that $i< j$ for any $j\in \A^{\Null}\setminus \{i\}$, otherwise, we can conclude that $\A^{\Null}\cap \tau^*=\emptyset$~.
\end{lemma}

\begin{proof}
Note that if $\A^{\Null}\cap \tau^*\neq \emptyset$ there is at least one element with rank $r$ in $\A^{\Null}$. For this element we may apply \cref{lemma:gapbound}. Note that this element is unique. If  $\A^{\Null}\cap \tau^*= \emptyset$, then the choice probability of all the elements in $\A^{\Null}$  are equal to $1/(r+1)$. 

Assuming $w_{\min}\geq \log 3/\beta$, we may bound the right-hand side of \cref{lemma:gapbound} as follows:

Since $a_r\in \tau^*$
\begin{align*}
 1-\exp(-\beta w_{a_r})\geq 1-\exp(-\beta w_{\min})&\geq 1-\exp(\log 3)\\
 &\geq 2/3~.   
\end{align*}

Therefore,

\begin{align*}
    \frac{1-\exp(\beta w_{a_r})}{1+r\exp(\beta w_{a_r})}&=  \frac{1-\exp(-\beta w_{a_r})}{1-r\left(1-\exp(-\beta w_{a_r})\right)+r}
\end{align*}

We substitute $x=1-\exp(-\beta w_{a_r})$, note $1\geq x\geq 2/3$.
\begin{align*}
    \frac{1-\exp(-\beta w_{a_r})}{1+r\exp(-\beta w_{a_r})}&=  \frac{x}{1+r-rx}
\end{align*}

Note that we have:
$\frac{1+r}{1+2r}\leq 2/3$ for $r\geq 1$. 

Thus, 
\begin{align*}
    \frac{x}{1+r-rx}\geq \frac{1}{1+r}&\iff x(1+r)\geq 1+r-rx\\
    & \iff x+rx\geq 1+r -rx\\
    &\iff x(1+2r)\geq 1+r\\
    & \iff x\geq (1+r)/(1+2r)\\
    &\Leftarrow x\geq 2/3~. 
\end{align*}

We now consider random variables $Y_{ij}$ for any $i,j=1,2,\dots,n$ as defined before.

We have:

\begin{equation}\label{eq:twocases}
  \mathbb{E}[Y_{ij}]
\begin{cases}
  \geq 1/(r+1) & \text{ if } ~ i=a_r \\ 
  = 0 & \text{ if } ~ i,j\in \bar{\tau^*}
\end{cases}
\end{equation}

Using Hoeffding bound, we have:

\begin{align*}
 \mathbb{P}\left(
\left| Y_{ij}-\mathbb{E}[Y_{ij}]\right| \geq 1/2(1+r)
\right)&\leq 2 \exp\left(- \frac{T}{32 (1+r)^2}  
\right)
\end{align*}

Which for  $T\geq 64\zeta (r+1)^2 \log (n+r)$ is bounded by: 

\begin{align*}
   2 \exp\left(- \frac{64\zeta (r+1)^2 \log (n+r) }{32 (1+r)^2}  
\right)&=2 \exp\left(-2\zeta\log (n+r)\right)\\
&= 2n^{-2\zeta}\cdot r^{-2\zeta} ~.
\end{align*}

In order to find $a_r$, we estimate $Y_{ij}$ for all $\Theta(r^2)$ pairs of $ij$ in $\A^{\Null}$. Using the above bound, and applying a union bound over all pairs we have:

With probability $1-2n^{-2\zeta}$:

\[
\forall i,j\in{\A^{\Null}}, ~ \left| Y_{ij}-\mathbb{E}[Y_{ij}]\right| \leq 1/2(1+r)
\]

Using \cref{eq:twocases}, this means that:

With probability $1-2n^{-2\zeta}$:
\begin{align*}
    \forall j\in{\A^{\Null}}, ~  Y_{{a_r}j}>  1/2(1+r),     \text{ and }\\
 \forall i,j\in \bar{\tau^*}~, Y_{ij}< 1/2(1+r)
\end{align*}

We use the above rule to find $a_r$: first we find all $Y_{ij}$s, and output the $i$ for which we have:

\begin{equation}\label{eq:findtop}
    \forall j\in{\A^{\Null}}\setminus \{i\},  ~  Y_{{i}j}>  1/2(1+r)~.
\end{equation}

Note that if $a_r$ exists, there will be one unique $i\in \A^{\Null}$ 
\cref{eq:findtop} holds. 
Therefore, we take $a_r=i$. 
Alternatively, if there is no element with rank $r$ we have  $\forall i,j\in \A^{\Null}, i,j\in \bar{\tau^*}$ thus, $Y_{ij}< 1/2(1+r)$.
\end{proof}

\begin{theorem}[Restatement of \cref{thm:learn2}]
Assume that $\beta\geq \log 3/w_{\min}$ and $\Null\notin \tau^*$. By only receiving $N$ and $r$ as input and being able to collect choice samples $\mathbf{D}$ adaptively, with probability at least $1-o(1)$, \BuCC\ is able to learn $\tau^*$ and $k$ from $\mathbf{D}$ using only $\Theta(r^2\log n)$ choice samples. 
\end{theorem}
\begin{proof}
Note that \BuCC\ calls \FindTop\ at most $\Theta(k+n/r+k^2\log_r k)=  \Theta(n/r+k^2\log_r k)$ times.  
    We take  $\zeta=3$, and
    using  \cref{lem:findtopanalysis} we have that by $\Theta\left(3 (r+1)^2\ \log n \right)$ choice samples which are obtained from presentation of $\A$ with probability $1-o(n^{-3})$  we will be able to identity the top element of $\A$. Since the total number of pairs $i,j$ is $\Theta(n/r+ k^2\log k)=\O(n^2\log n)$, we need to take a union bound over $n^2\log n$ variables thus, in total the probability of failure is bounded by $o(n^2n^{-3})=o(1)$, from which we conclude the premise.

\end{proof}

\section{Experimental setup and extra tables and figures}\label{app:ex}

\subsection{Details of learning choice probabilities  on the Sushi data-set. }

As explained in the main body, we have divided the data set into two non-intersecting parts for training (80\%) and test sets (20\%). 
We use the training set to learn choice probabilities. First we employ \BuCC-II (\cref{alg:bucchoi}). \BuCC-II uses the top-$10$ sushi types as samples generated from an unknown distribution and collects choice data from it by actively selecting assortments.  Since the result of \BuCC-II may produce partial orders, we break the ties, by using a score function based on the random variables $X_{ij}$ that are used in \FindTop. In particular, since $X_{ij}$ shows the number of times item $i$ beats $j$ we use $\sum_{j\in N}X_{ij}$ as a tie breaking score for item $i$.  

This is needed because we need to feed the learned center from \BuCC-II to \DeepCheep\ to learn choice probabilities and \DeepCheep\ receives as input  a top-$k$ list. 
Note that $k$ is unknown to the algorithm and we let \BuCC\ learn it. If $k$ is too large (>15) we truncate the learned center to reduce time complexity.

\begin{algorithm}
\caption{\BuCC -II}\label{alg:bucchoi}
\begin{algorithmic}
\STATE \textbf{Input}, $N$: set of products, $m$: number of samples, choice  oracle $\choice_\tau;~ \tau\sim \D$ where $\D$ is a topKGMM. 
\STATE \textbf{Output} center of $\D$
\STATE $T=\emptyset$
\FOR{$i\in N$}
\STATE $S=\emptyset$
\STATE $\A=\{i\}$
\FOR{$j=1:m$}
\STATE $c_j=\choice_\tau(\A)$
\STATE $S=S\cup \{c_j\}$
\ENDFOR
\STATE $T=T\cup \FindTop(\A, S)$
\ENDFOR
\FOR{$i,j\in T$}
\STATE $\A=\{i,j\}$
\FOR{$j=1:m$}
\STATE $c_j=\choice_\tau(\A)$
\STATE $S=S\cup \{c_j\}$
\ENDFOR
\IF{$i==\FindTop(\A,S)$}
\STATE $\sigma[i]>\sigma[j]$
\ELSE 
\STATE $\sigma[i]<\sigma[j]$
\ENDIF
\ENDFOR
\STATE \textbf{Return } $\sigma$
\end{algorithmic}
\end{algorithm}

We tune parameters $\beta$ and $p$ by performing a grid search over a range of values as in \Cref{table:allparams_onecluster}. From this table we conclude that $p=0.5$ and $\beta=0.05$ have the best test error. Thus, we select them as the model parameters. The mean and std of test error of the best parameters are reported in \cref{fig:test:nocl}.

We handle learning mixture models by clustering the data into several clusters. First we use $\K^p$   as a distance metric taking $p\in [0.01,0.25,0.05,0.75,0.1,0.25,0.5,1,1.5,2.2.5,5]$ and then we divide data set to $2,3$ or $5$ clusters. Most of the clusters we obtain have a negative silhouette coefficient, and the only positive ones are:

p= 0.1 num clusters = 2: silhouette score = 0.006287393358176328\\
p= 1.25 num clusters = 2:
silhouette score = 0.007796245381136862\\
p= 2.5 num clusters = 2: silhouette score = 0.004498089502999035\\
p= 5 num clusters= 2: silhouette score = 0.011002010781645042\\ 
p= 5 num clusters = 3:
silhouette score = 0.0023222431733978116

As we can see, even when the silhouette scores are positive they are pretty low, which suggests that one cluster is the best choice. This is consistent with the test error we obtained, as the model trained without clustering has a lower test error compared to the model trained on two clusters or more; see \cref{fig:test:twocl}.

\newcommand{\highlightcell}[1]{%
  \tikz[baseline=(X.base)] \node[draw=black, rectangle, rounded corners=2pt, inner sep=2pt] (X) {#1};%
}

\begin{table}[h]
\centering
\small
\begin{tabular}{|c|cccccccccc|}
\hline 
\diagbox{$p$}{$\beta$}&0.05&0.1&0.25&0.5&0.75&1&1.25&1.5&1.75&2\\
\hline 
0.01&
0.1162& 0.1047&  0.0888& \cellcolor{pink}0.0879& 0.0925&
 0.0979& 0.1030& 0.1087&0.1145& 0.1202 
\\
0.025
&0.1133& 0.1007& \cellcolor{pink}0.0858& 0.0872& 0.0922&
 0.09768& 0.1028& 0.1085& 0.1144&  0.1201
\\  
0.05& 
0.1083& 0.0938&\cellcolor{pink} 0.0817&0.0852& 0.0907&
 0.0964& 0.1017& 0.1075&0.1135& 0.1193
\\
0.075& 0.1032& 0.0868&\cellcolor{pink} 0.0797& 0.0841& 0.0898&
 0.0956& 0.1011& 0.1070& 0.1130& 0.1188\\ 
0.1 &0.0980& 0.0798& \cellcolor{pink}0.0768&0.0819& 0.0880&
 0.0941& 0.0997& 0.1058& 0.1119& 0.1178\\
 0.25&
0.0709&\highlightcell{\cellcolor{pink} 0.0598}& 0.0658& 0.0732& 0.0808&
 0.0879& 0.09433& 0.1009& 0.1075&   0.1139
\\
0.5& \highlightcell{\cellcolor{pink}0.0445}& 0.0555& 0.0629& 0.07098& 0.0789&
 0.0863& 0.0929& 0.0997& 0.1064& 0.1128\\
1&\highlightcell{\cellcolor{pink} 0.0504 }&0.0619 &0.0677&0.0748& 0.0821
& 0.0890  &0.0953&  0.1018 &0.1083& 0.1146\\
1.5& \cellcolor{pink}0.0696& 0.07479& 0.0773&0.0825& 0.0885&
 0.09454& 0.1001& 0.1061& 0.1122& 0.1181\\  
2& \cellcolor{pink}
0.0791&  0.08115& 0.0821&0.0863& 0.0917&
 0.0972&  0.1025&0.1082&  0.1141& 0.1198
\\
2.5&\cellcolor{pink} 0.0825& 0.0834& 0.0838& 0.08769& 0.0928&
 0.09824& 0.1033& 0.1090& 0.1148& 0.1204\\
5&\cellcolor{pink} 0.0851& 0.0851& 0.0851& 0.0887& 0.0937&
 0.0989& 0.1039&0.1095& 0.1153&  0.1209\\
\hline 
\end{tabular}
\caption{\footnotesize{Average tests errors for several choices of $p$ and $\beta$ used for parameter tuning. No clustering has been performed. The average test error we obtain for MNL is 0.168.}}\label{table:allparams_onecluster}
\end{table}

\begin{table}[h]
\centering
\small
\begin{tabular}{|c|ccccccc|c|}
\hline 
& $\beta=0.05$& $0.1$& $0.25$&  $0.5$& $0.75$& $1$& $1.25$& MNL\\
\hline 
  $p=0.1$& 
  0.0997&0.0771&\cellcolor{pink} 0.0732 &\cellcolor{pink} 0.0748 &0.0785&
 0.083 &0.0876&0.1877\\
\hline 
 $p=1.25$ & 
\cellcolor{pink} 0.0788&0.09273& 0.1022& 0.1116&0.1224&
 0.1324& 0.1409&0.1883\\
\hline 
\end{tabular}
\caption{\footnotesize{average test error of learning choice probabilities, comparison of Mallows model with different $\beta$s and MNL;  2 clusters are generated based on distance function $ {\K}^{p}$ for given $p$. The best values are highlighted.  }}\label{table:testerror-allbetas}
\end{table}

\subsection{Runtime of \PRIM\ and \DeepCheep}\label{sec:app:runtimes}

\begin{table}[H]
\centering
\small
\begin{tabular}{|l|ccccc|ccccc|}
\hline 
& \multicolumn{5} {c|} {pre-processing time} & \multicolumn{5} {c|} {amortized time per sample}\\
&$k=8$&$k=10$&$k=12$&$k=14$&$k=16$&$k=8$&$k=10$&$k=12$&$k=14$&$k=16$\\
\hline 
$n=200$&0.007& 0.035& 0.19&1.06& 8.64&5.67e-05&9.59e-05&0.00022&0.00074& 0.0032\\
$n=500$&0.008&0.044&0.23&1.24&7.83&7.89e-05&0.00012&0.00025&0.00076&0.0028\\
$n=1000$&0.012&0.079&0.29&1.46&8.39&0.00011& 0.00022&0.00031& 0.00082&0.0030\\

\hline 
\end{tabular}
\caption{\footnotesize{average (among 10 runs) runtime of sampling algorithm \PRIM\ in seconds, $\beta=0.2$}}\label{tab:prim-runtimes}
\end{table}

\begin{figure}[H]
    \centering
\includegraphics[width=0.5\linewidth]{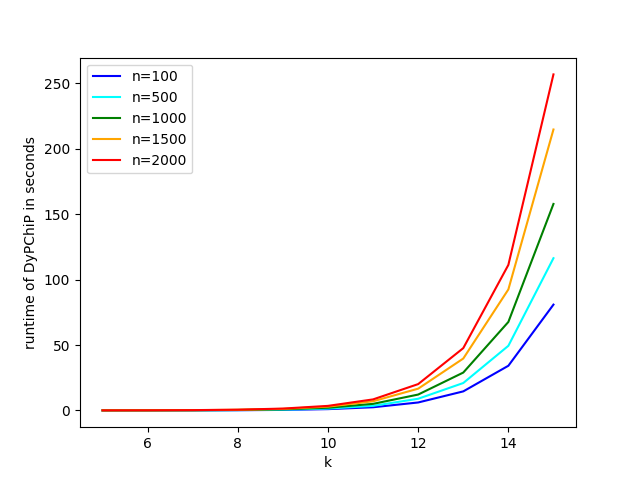}
    \caption{Runtime of \DeepCheep\ (in seconds) for various choices of $n,k$. Here $\beta=0.6$, $p=0.5$ and $w=2\vec{1}$, size of assortment $r=k-2$.}
    \label{fig:dypchipruntimes}
\end{figure}

\subsection{Additional tables and figures for the analysis of sample complexity of \BuCC\ and \FindTop\ on synthetic data}\label{sec:app:learningalgs}

\begin{table}[h]
\centering
\small
\begin{tabular}{|l|cc|}
\hline 
&$k=6$&$k=10$\\
\hline 
$n=1000$&$6.95\pm 5.06$&$0.4\pm 0.48$\\
$n=5000$&$228.5\pm53.3$&$228.6\pm 64.56$\\
$n=10000$&$934.05\pm 209.35$&$870.1\pm 259.64$\\
$n=20000$&$3505.05\pm 409.87$&
$3282.55\pm 708.9$\\
\hline 
\end{tabular}
\quad 
\begin{tabular}{|l|cc|}
\hline 
&$k=6$&$k=10$\\
\hline 
$n=1000$&$0.1\pm 0.3$&$0.0\pm 0.0$\\
$n=5000$&$0.35\pm 0.50$&$1.92\pm 0.92$\\
$n=10000$&$2.95\pm 2.28$&$3.0\pm 3.16$\\
$n=20000$&$5.1\pm3.5$& 
$9.0\pm13.2$
\\
\hline 
\end{tabular}
\caption{\footnotesize{Kendall's Tau distance of true and learned center by \BuCC\  with $100$ (left table) or $200$ (right table) samples for  $n=1000,5000,10000,20000$ and $k=6,10$. We let $\beta=0.6$. distance reported as $mean\pm std$ these numbers are taken from 10 runs of the algorithm. }}
\end{table}

\begin{table}[h]
\centering
\small
\begin{tabular}{|l|ccccc|}
\hline 
&$\beta=0.4$&$\beta=0.6$&$\beta=0.8$&$\beta=1$&$\beta=1.2$\\
\hline 
50 samples&$12.75\pm 4.4$&$8.25\pm 7.7$&$7.05\pm 5.64$& $4.35\pm 7.5$ & $0.7\pm 1.65$\\
\hline 
100 samples &$3.5\pm 0.67$&$0.35\pm 0.549$& $0.0\pm 0.0$& $0.0\pm 0.0$& $0.0\pm0.0$
\\
\hline 
\end{tabular}
\caption{\footnotesize{Kendall's Tau distance of true and learned center  by \BuCC\ with $50$ and $100$ samples for various  $\beta$, $n=1000$, $k=12$, $p=0.5$ reported as $mean\pm std$. mean and standard deviation is taken for 10 runs of the algorithm for each set of parameters.  }}
\end{table}

\begin{figure}
    \centering
      \begin{subfigure}[b]{0.40\linewidth}
        \centering
  \includegraphics[width=\linewidth]{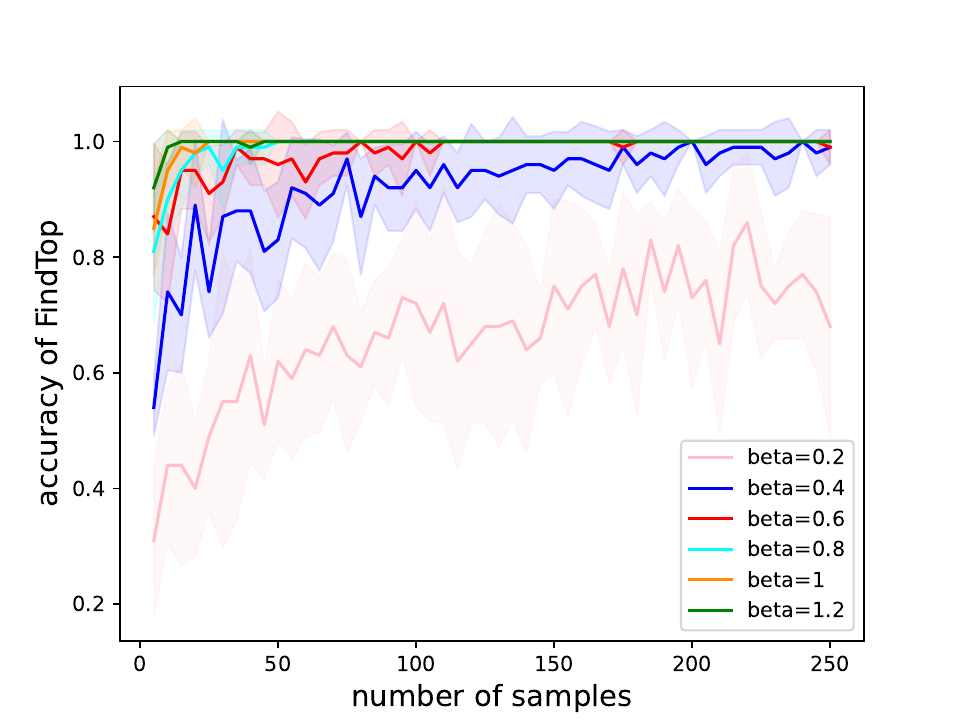}
        \caption{\tiny{$n=1000, k=12, r=10,p=0.5,  w=2\vec{1}$.  }}
    \end{subfigure}
        \begin{subfigure}[b]{0.40\linewidth}
        \centering
  \includegraphics[width=\linewidth]{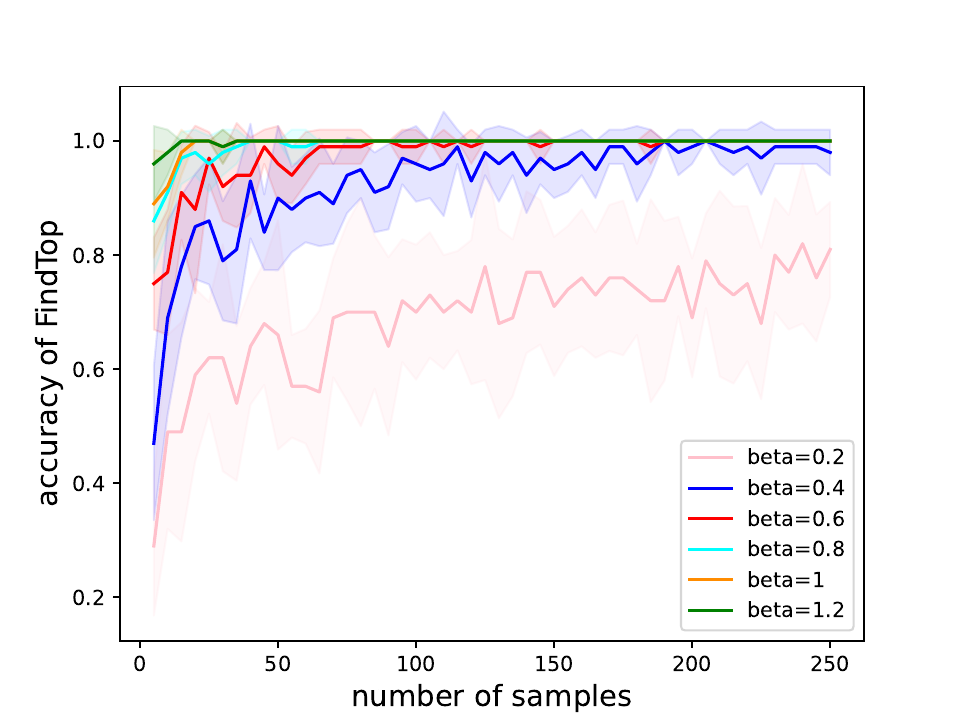}
        \caption{\tiny{$n=1000, k=14, r=12,p=0.5,  w=2\vec{1}$.  }}
    \end{subfigure}
        \begin{subfigure}[b]{0.40\linewidth}
        \centering
  \includegraphics[width=\linewidth]{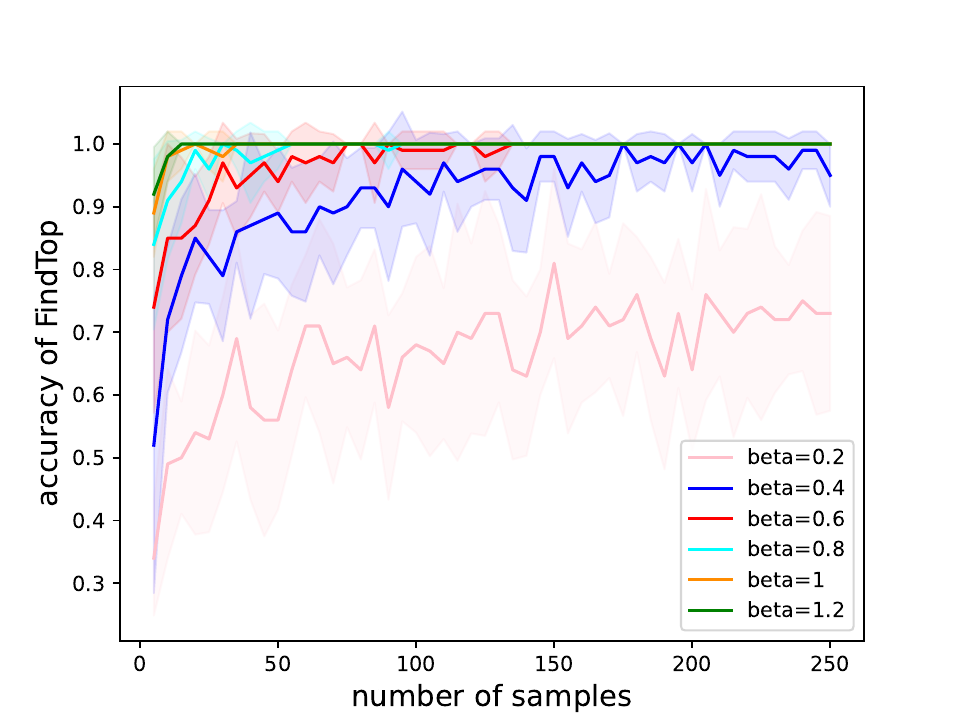}
        \caption{\tiny{$n=500, k=12, r=10,p=0.5,  w=2\vec{1}$.  }}
    \end{subfigure}
      \begin{subfigure}[b]{0.40\linewidth}
        \centering
\includegraphics[width=\linewidth]{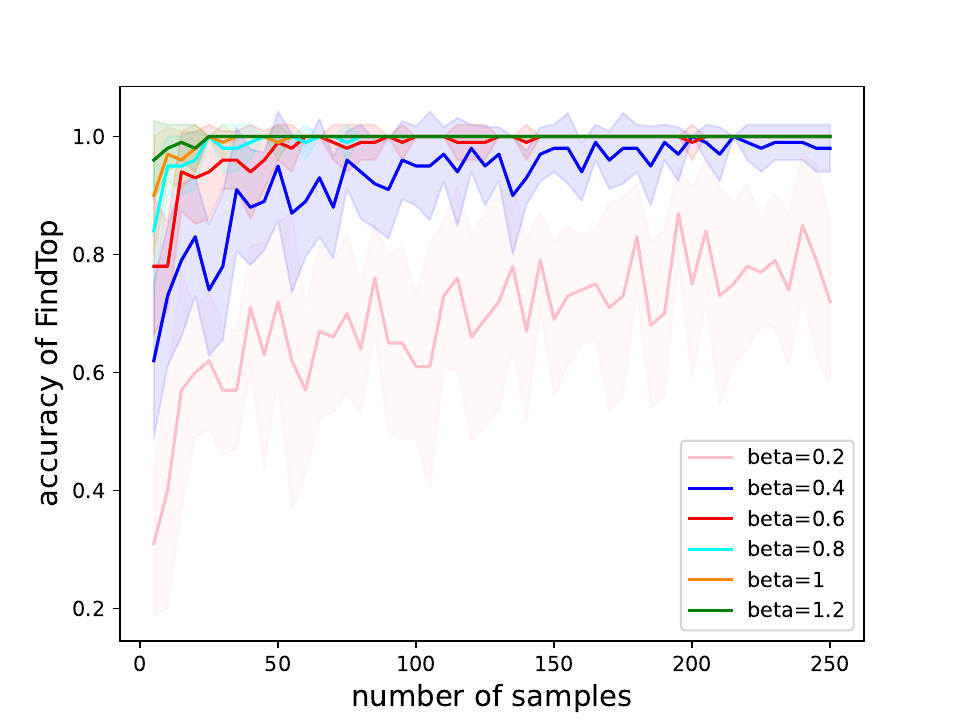}
        \caption{\tiny{$n=500, k=14, r=12,p=0.5,  w=2\vec{1}$.  }}
    \end{subfigure}

    \begin{subfigure}[b]{0.40\linewidth}
        \centering
\includegraphics[width=\linewidth]{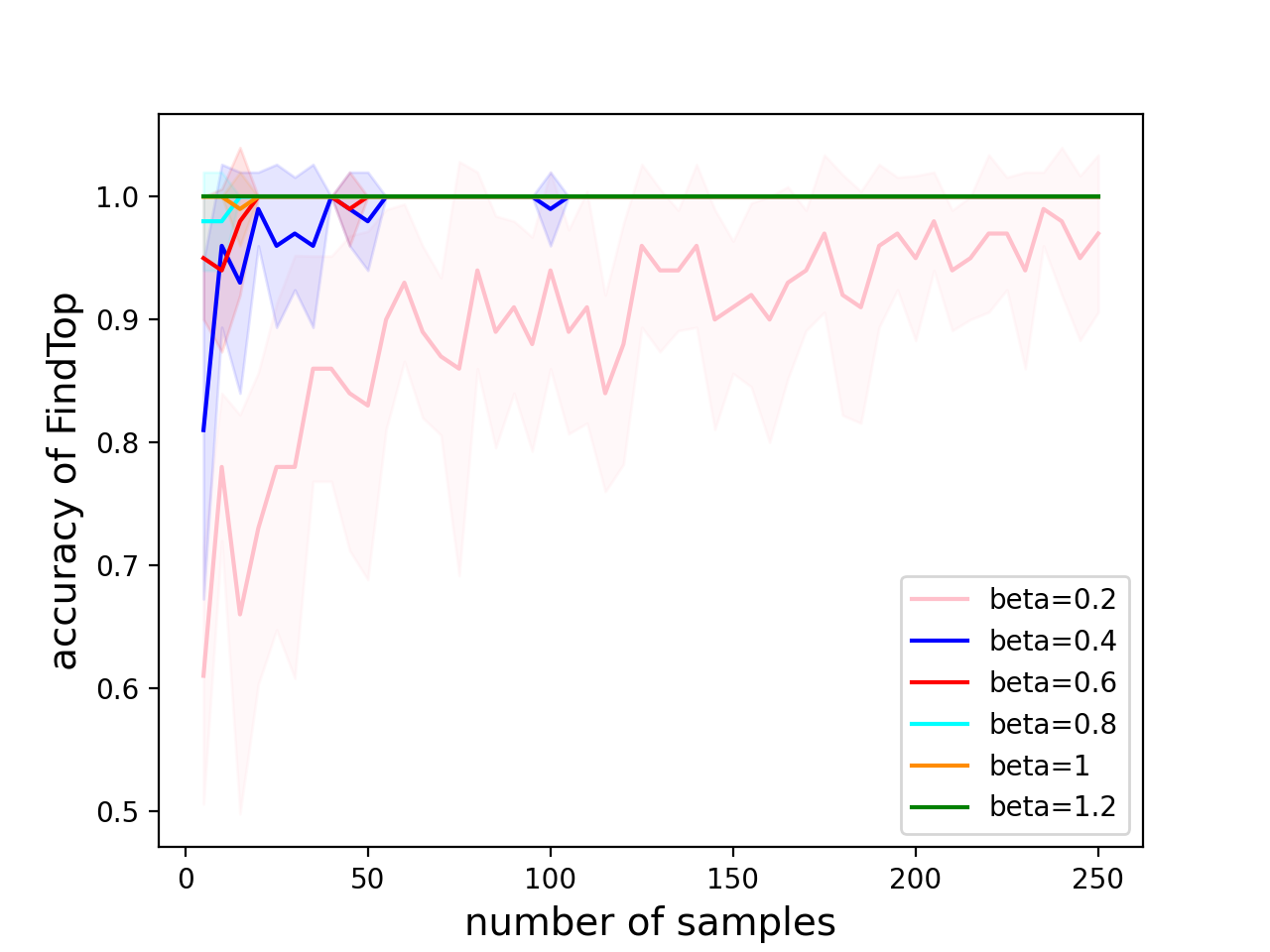}
        \caption{\tiny{$n=300, k=8, r=6,p=1.5,  w=322\vec{1}$.  }}
    \end{subfigure}
      \begin{subfigure}[b]{0.40\linewidth}
        \centering
\includegraphics[width=\linewidth]{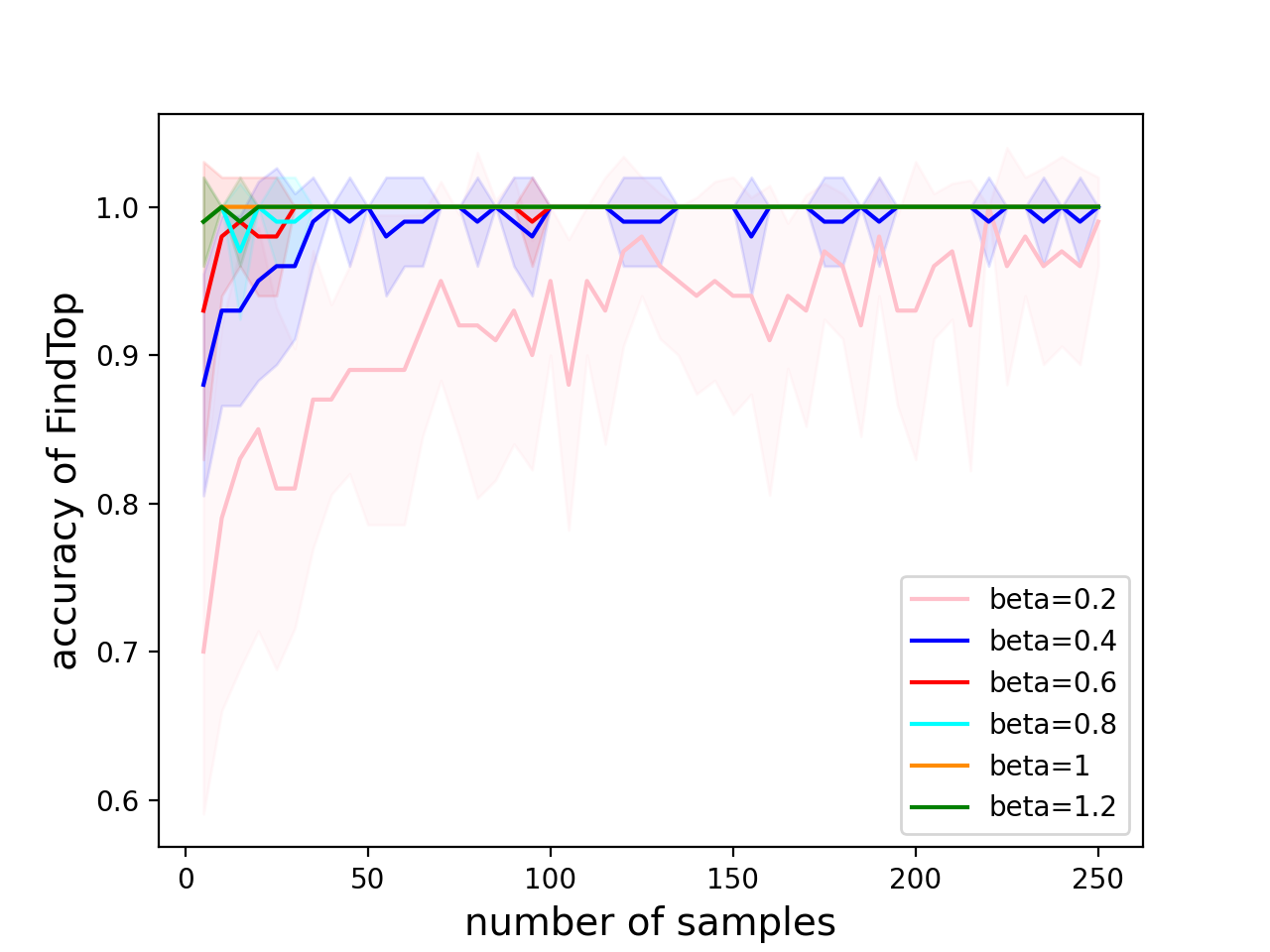}
        \caption{\tiny{$n=300, k=8, r=5,p=2,  w=32222211$.  }}
    \end{subfigure}
    \caption{Sample complexity of \FindTop\ for a wide range of parameters.}
    \label{fig:findTop_allfigs
    }
\end{figure}

\begin{figure}
    \centering
      \begin{subfigure}[b]{0.40\linewidth}
        \centering
  \includegraphics[width=\linewidth]{Source/figures-tables/BuCCAcc_n1000_k10_r8_p0.5_bigfont_elevenvals.png}
        \caption{\tiny{$n=1000, k=10, p=0.5,  w=2\vec{1}$.  }}
    \end{subfigure}\quad 
        \begin{subfigure}[b]{0.40\linewidth}
        \centering
  \includegraphics[width=\linewidth]{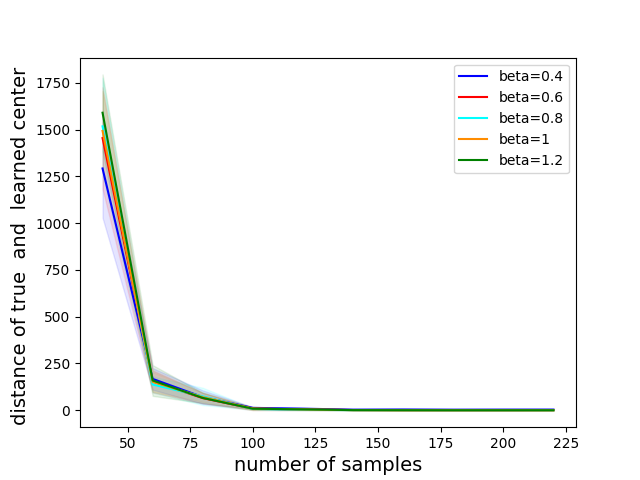}
        \caption{\tiny{$n=1000, k=14, p=0.5,  w=2\vec{1}$.  }}
    \end{subfigure}
      \begin{subfigure}[b]{0.40\linewidth}
        \centering
  \includegraphics[width=\linewidth]{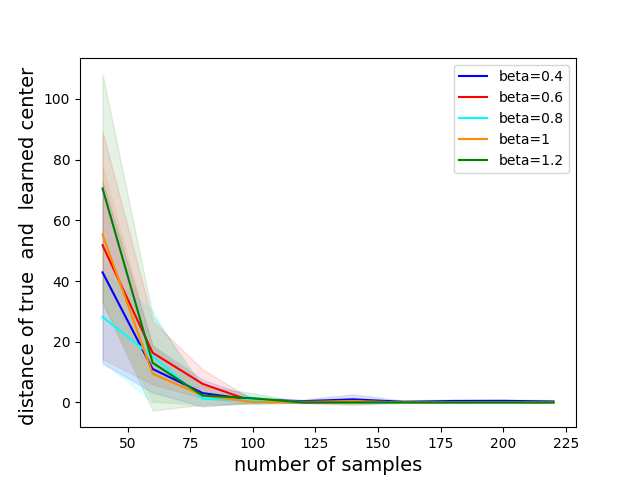}
        \caption{\tiny{$n=300, k=8, p=1.5,  w=322\vec{1}$.  }}
    \end{subfigure}
         \begin{subfigure}[b]{0.40\linewidth}
        \centering
  \includegraphics[width=\linewidth]{Source/figures-tables/BuCCAcc_n300_k8_r5_p2_bigfont_elevenvals.png}
        \caption{\tiny{$n=300, k=8, p=2,  w=32222111$.  }}
    \end{subfigure}
    \caption{Sample complexity of \BuCC\ for a wide range of parameters. y axis shows the Kendall's Tau distance between true and learned center while x axis shows sample complexity }
    \label{fig:BuCChoi_allparams
    }
\end{figure}

\end{document}